\documentclass[3p,sort & compress]{elsarticle}
\usepackage{amssymb}
\usepackage{latexsym}
\usepackage{amsfonts}
\usepackage{amsthm}
\usepackage{amsbsy}
\usepackage{amsmath}
\usepackage{graphics,graphicx}
\usepackage{subfigure}
\usepackage{multirow,multicol}
\usepackage{bm}
\usepackage{bbm}
\usepackage{diagbox}
\usepackage{color}
\usepackage{float}
\usepackage{calc}
\usepackage{caption}
\usepackage{dsfont}
\usepackage{ifthen}
\usepackage{mathrsfs}
\usepackage[colorlinks,linkcolor=blue,citecolor=blue]{hyperref}
\usepackage{epstopdf}
\usepackage{epsfig}
\usepackage{algorithm}
\usepackage{algorithmic}
\usepackage{pifont}
\usepackage{natbib}
\usepackage{overpic}
\usepackage{psfrag}
\usepackage{rotating}
\usepackage{stmaryrd}
\usepackage{verbatim}
\usepackage{setspace}
\usepackage{tikz}
\usepackage{booktabs}
\usepackage{comment}
\usepackage{listings} 
\usepackage{xcolor} 
\newdefinition{remark}{Remark}
\newdefinition{method}{Method}
\newdefinition{example}{Example}

\numberwithin{equation}{section}

\numberwithin{theorem}{section}

\newcommand{\orcid}[1]{\href{https://orcid.org/#1}{\includegraphics[width=8pt]{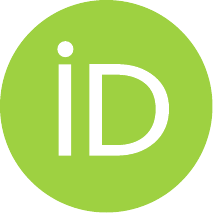}}}
\journal{Journal of \LaTeX\ Templates}
\begin{document}
\begin{frontmatter}
    \title{Enhanced BPINN Training Convergence in Solving General and Multi-scale Elliptic PDEs with Noise}
    \author[Berkeley]{Yilong Hou\fnref{fn1}}\ead{yilong$\_$hou@berkeley.edu}
    \author[Ceyear,SDU]{Xi'an Li\orcid{0000-0002-1509-9328}\corref{equalcon},\corref{cor1}\fnref{fn1}}\ead{lixian9131@163.com}
    \author[uq]{Jinran Wu\orcid{0000-0002-2388-3614}} \ead{wujrtudou@gmail.com}
    \author[GUFE]{You-Gan Wang\orcid{0000-0003-0901-4671}\corref{cor1}}\ead{ygwang2012@gmail.com}
    \cortext[cor1]{Corresponding author.}
    \fntext[equalcon]{These authors contributed equally to this work.}
    \address[Berkeley]{University of California at Berkeley, CA, USA}
    \address[Ceyear]{Ceyear Technologies Co., Ltd, Qingdao 266555, China}
    \address[SDU]{Shandong University, Qingdao 266237, China}
    \address[uq]{The University of Queensland, St Lucia 4072, Australia}
    \address[GUFE]{Guangdong University of Finance and Economics, Guangzhou 510320, China}
    
    \begin{abstract}
    Bayesian Physics Informed Neural Networks (BPINN) have attracted considerable attention for inferring the system states and physical parameters of differential equations according to noisy observations. However, in practice, Hamiltonian Monte Carlo (HMC) used to estimate the internal parameters of the solver for BPINN often encounters these troubles including poor performance and awful convergence for a given step size used to adjust the momentum of those parameters. To address the convergence of HMC  for the BPINN method and extend its application scope to multi-scale partial differential equations (PDE), we develop a robust multi-scale BPINN (dubbed MBPINN) method by integrating multi-scale deep neural networks (MscaleDNN) and the BPINN framework. In this newly proposed MBPINN method, we reframe HMC with Stochastic Gradient Descent (SGD) to ensure the most ``likely'' estimation is always provided, and we configure its solver as a Fourier feature mapping-induced MscaleDNN. This novel method offers several key advantages: (1) it is more robust than HMC, (2) it incurs less computational cost than HMC, and (3) it is more flexible for complex problems. We demonstrate the applicability and performance of the proposed method through some general Poisson and multi-scale elliptic problems in one and two-dimensional Euclidean spaces. Our findings indicate that the proposed method can avoid HMC failures and provide valid results. Additionally, our method is capable of handling complex elliptic PDE and producing comparable results for general elliptic PDE under the case of lower signal-to-noise rate. These findings suggest that our proposed approach has great potential for physics-informed machine learning for parameter estimation and solution recovery in the case of ill-posed problems.
    \end{abstract}
\end{frontmatter}

\section{Introduction}\label{sec:1}
Partial differential equations (PDE) have extensive applications across various fields, as they model the physical properties and behaviors of complex systems. The parameters within these PDEs often represent key physical properties of the system under study~\cite{pensoneault2024efficient}. To estimate these parameters or recover the solution based on observations and limited constraints, which can provide insights into the underlying physical phenomena \cite{engl1996regularization}. In the context of well-posed problems, physics-informed neural networks (PINN) have demonstrated significant success in accurately estimating these parameters. However, in ill-posed problems, where the observations may be noisy or incomplete, the traditional numerical solvers often fail to provide reliable solutions \cite{engl1996regularization}. To address these challenges, statistical and machine learning tools, such as regularizers, are employed to handle the ill-posed inverse problems by selecting or weighting variables \cite{beck1977parameter, woodbury2002inverse}. Bayesian statistics, as a natural regularization approach, has found numerous applications in dealing with noisy and high-dimensional data \cite{wang2004hierarchical}. By integrating Bayesian statistics with PINN, some researchers have proposed robust methodologies for estimating parameters from real-world observational data in linear or nonlinear systems, thereby enhancing the accuracy and reliability of the analysis \cite{yang2021b, yang2022multi,linka2022bayesian}.

\subsection{Bayesian PINN}\label{subsec:WhyBPINN}
In terms of the solution for the inverse problem governed by PDE, to improve the capacity of PINN for separating the system states and the noise from real noisy observation, we need to estimate well the internal parameters of PINN and system parameters based on those observations. Bayesian statistics offers effective inference methods for noisy and incomplete data \cite{li2014adaptive}. By treating the parameters of PINN as random variables instead of unknown deterministic values, the Bayesian approach ultimately provides a distribution conditioned on observations as the inference of these parameters. When seeking a point estimate, such as the estimation of PINN parameters, Bayesian methodology regularizes the parameters so that useful parameters receive more weight, thereby controlling the error \cite{wang2004hierarchical}. 

Consequently, Bayesian PINN is proposed using a Bayesian framework to estimate PINN parameters. Considering noise sampled from mean zero, i.i.d. normal distributions, BPINN uses Bayesian statistics as the estimation method and treats the likelihood accordingly. Finally, it employs Hamiltonian Monte Carlo (HMC) or variational inference (VI) for posterior sampling. This framework successfully quantifies uncertainty and improves predictions in noisy environments.

\subsection{HMC and VI} \label{subsec:HMCVI}
The prior of BPINN can be assumed to be a Gaussian Process with mean 0 as the number of parameters goes to infinity \cite{neal2012bayesian, lee2017deep}, but the distribution of the posterior is generally complicated and non-Gaussian due to the nonlinear nature of neural networks. Since the posterior distribution has no analytical expressions for the mean or the variance, they could only be approximated. There are generally two ways of approximating the posterior distribution, which are the HMC method and the VI method. 

The HMC method is an enhancement of the Markov Chain Monte Carlo (MCMC) method. The new step of HMC is generated by solving the Hamiltonian system instead of a random walk in MCMC, but the acceptance procedure of each new step remains the same \cite{betancourt2019convergence, neal2011mcmc, betancourt2017conceptual}. HMC is more efficient than MCMC by higher acceptance rate, but the computational cost remains high \cite{pensoneault2024efficient}. 

VI assumes that the posterior distribution belongs to a family of parameterized distributions. In other words, VI uses functions with a different parameterization to approximate the posterior distribution. By updating the parameters from this different parameterization, VI minimizes the KL divergence between the posterior distribution and its approximation. VI considers the optimal solution of this deterministic optimization problem as the best approximation of the posterior distribution \cite{blei2017variational, yang2021b}

This paper chooses HMC over VI for theoretical and practical reasons. Theoretically, VI is projecting the posterior distribution onto a new function class which is usually assumed to be the mean-field Gaussian approximation by the deep learning community \cite{foong2019between, yao2019quality}. Therefore, VI does not offer the same theoretical guarantees as MCMC approaches. Practically, lots of papers show that HMC has better performance than VI. For example, in high-dimensional problems \cite{yang2021b} showing HMC's superiority over VI and challenges.

\subsection{Objective} \label{subsec:objective} 

Since the establishment of BPINN, they have demonstrated remarkable performance in solving mathematical problems in scientific computations and engineering applications based on their great potential in integrating prior knowledge with data-driven approaches. For example, utilizing BPINN to quantify uncertainties in the predictions of physical systems modeled by differential equations \cite{yao2019quality, ceccarelli2019bayesian}, inverse problems \cite{li2024bayesian, pensoneault2024efficient, antil2021novel}, nonlinear dynamical system \cite{linka2022bayesian}, etc.  

After that, many efforts have been made to enhance the performance of BPINN are concluded as two aspects:  the improvement of posterior sampling methods and the choice of NN-solver. In terms of the sampling strategy, one approach is to reduce the computational cost of the MCMC method \cite{lin2022multi, li2023surrogate}, and the other approach is to breach the gap between MCMC and VI through a partical-based VI approach. In the context of BPINN, methods like Stein Variational Gradient Descent (SVGD) \cite{sun2020physics} and Ensemble Kalman inversion (EKI) \cite{jiang2022pinn, iglesias2013ensemble, pensoneault2024efficient} has been proposed for sparse and noisy data. In terms of the latter one, the authors in \cite{yang2022multi} reconstructed the solver of PINN by extending the output pipelines and ensembled the multiple outputs at the same point to calculate statistical properties, then imposed any prior knowledge or assumptions regarding the uncertainty of the data. A Generative Adversarial Networks model is configured as the solver of BPINN (BPI-GAN) to learn flexible low-dimensional functional priors, e.g., both Gaussian and non-Gaussian processes, then BPI-GAN is easy to apply to big data problems by enabling mini-batch training using stochastic HMC or normalizing flows \cite{meng2022learning}. To robustly address multi-objective and multiscale problems, a novel methodology for automatic adaptive weighting of Bayesian Physics-Informed Neural Networks, which automatically tunes the weights by considering the
multitask nature of target posterior distribution\cite{perez2023adaptive}.

In practice, we found out that HMC is sensitive to the step size of updating parameters and sometimes does not converge to a stable distribution. The step size is used to adjust the momentum in the Hamiltonian system. When this step size is too large, the sum of the log-likelihood of the parameters goes to infinity and the algorithm breaks down. However, there are no explicit standards for the measure of step size. That is to say, one certain step size may work for a PDE, but fail for a different PDE. And even for the same PDE, changing the step size could make a huge impact on the performance of BPINN.

Additionally, as in the aforementioned BPINN methods, the solver is configured as a vanilla deep neural networks (DNN) model, then their performance will be limited by the spectral bias (or called frequency preference) of DNN and they may encounter some dilemmas for addressing complex problems, such as recovering the solution of multi-scale PDEs from noisy data \cite{Xu_2020,rahaman2018spectral}. Recently, a multi-scale DNN (MscaleDNN) framework has been developed to resolve the issue of DNN which easily captures the low-frequency component of target functions but hardly matches the high-frequency component. Furthermore, a rondom Fourier feature embedding consisting of sine and cosine is introduced to improve the capacity of DNN, it will mitigate the pathology of spectral bias and enable networks to learn high frequencies more effectively\cite{rahaman2018spectral,wang2020eigenvector, tancik2020fourier,li2023deep}. 

In this paper, we develop an enhanced BPINN method by embracing the multi-scale DNN into BPINN framework (dubbed MBPINN), and reconstruct its estimation method to address the convergence problems of HMC-driven BPINN (BPINN\_HMC) by replacing the original HMC with Stochastic Gradient Descent (SGD). Compared to the classical BPINN\_HMC, this new approach ensures point estimation of the parameters with significantly lower computational costs. Additionally, the Fourier feature mapping consisting is used to further improve the performance of MBPINN to address general and complex problems under different noise levels. We apply this method across various types of PDE problems, including linear and nonlinear Poisson equations as well as multi-scale elliptic PDEs.  The newly proposed method demonstrates strong potential and robustness in the aforementioned scenarios.

The remaining parts of our work are organized as follows. 
In Section \ref{sec:02}, we briefly introduce the formulation BPINN and its failure through an example of a multi-scale PDE.
Section \ref{sec:03} provides our solutions for both of the problems mentioned in the last section, by introducing the methodology of MBPINN and reframing HMC with SGD. 
In Section \ref{sec:04}, some scenarios of multi-scale and general Poisson PDEs are performed to evaluate the feasibility and effectiveness for  our proposed method. 
Finally, some conclusions of this paper are made in Section \ref{sec:conclusion}.

\section{Formulation and Failure of Classical BPINN }\label{sec:02}
\subsection{Formulation of BPINN } \label{subsec:BPINN}
In this subsection, we briefly introduce the formulation of BPINNs first proposed in \cite{yang2021b}. Given a $d$-dimensional domain $\Omega$ and its boundary $\partial \Omega$, let us consider the following system of parametrized PDEs:
\begin{equation}\label{eq2PPDE}
\begin{aligned}
&\mathcal{N}_{\bm{\lambda}}[u(\bm{x})]=f(\bm{x}), ~\quad \bm{x} \in \Omega \\
&\mathcal{B}[u\left(\bm{x}\right)]=g(\bm{x}), \quad\quad\bm{x} \in \partial \Omega\\
\end{aligned}
\end{equation}
in which $\mathcal{N}_{\bm{\lambda}}$ stands for the linear or nonlinear differential operator with parameters $\bm{\lambda}$, $\mathcal{B}$ is the boundary operators. Generally, the sampling data of force term $f(\bm{x})$ and boundary function $g(\bm{x})$ may be disturbed by unanticipated noise for real applications. The available dataset $\mathcal{D}$ composed by the collocation points and the corresponding evaluation of $f$ and $g$ for this scenario is given by
\begin{equation}
\mathcal{D} = \mathcal{D}_f \cup \mathcal{D}_g
\end{equation}
with 
\begin{equation*}
\mathcal{D}_f = \left\{\left(\bm{x}_f^{(i)},\tilde{f}^{(i)}\right)\right\}_{i=1}^{N_f}~~\text{and}~~\mathcal{D}_g = \left\{\left(\bm{x}_g^{(j)}, \tilde{g}^{(j)}\right)\right\}_{j=1}^{N_g}.
\end{equation*}
In which, the above observations are i.i.d Gaussian random variables, i.e.,
\begin{equation}
    \begin{aligned}
        \tilde{f}^{(i)} &= f(\bm{x}_f^{(i)}) + \varepsilon_f^{(i)}, \quad i = 1, 2, \ldots, N_f, \\
    \tilde{g}^{(j)} &= g(\bm{x}_g^{(j)}) + \varepsilon_g^{(j)}, \quad j = 1, 2, \ldots, N_g,
    \end{aligned}
\end{equation}
where $\bm{\varepsilon}_f = \{\varepsilon_f^{(1)}, \varepsilon_f^{(2)}, \cdots\}$ and $\bm{\varepsilon}_g = \{\varepsilon_g^{(1)},\varepsilon_g^{(2)},\cdots\}$ are independent mean-zero Gaussian noise with given standard deviations $\tau_f$ and $\tau_g$, respectively. Note that the size of the noise could be different among observations of different terms, and even between observations of the same terms in the PDE.

To recover the solution of \eqref{eq2PPDE} from noisy observations through the Bayesian approach, the BPINN starts from representing $u$ with a surrogate model $u_{NN}(\bm{x}; \bm{\theta})$, where $\bm{\theta}$ is the vector of parameters in the surrogate model with a prior distribution $P(\bm{\theta})$. When the process of the Bayesian method meets the physics-informed neural networks, the architecture of BPINN is constructed and its surrogate model is configured as a general DNN. Mathematically, the DNN defines the following mapping
\begin{equation}
   \mathcal{F}: \bm{x}\in\mathbb{R}^{d}\Longrightarrow \bm{y}=\mathcal{F}(x)\in\mathbb{R}^{c}
\end{equation}
with $d$ and $c$ being the dimensions of input and output, respectively. In fact, the DNN function $\mathcal{F}$ is a nested composition of sequential single linear functions and nonlinear activation functions, which is in the form of
\begin{equation}
	\begin{cases}
		\bm{y}^{[0]} = \bm{x}\\
		\bm{y}^{[\ell]} = \sigma\circ(\bm{W}^{[\ell]}\bm{y}^{[\ell-1]}+\bm{b}^{[\ell]}), ~~\text{for}~~\ell =1, 2, 3, \cdots\cdots, L
	\end{cases}
\end{equation}
where $\bm{W}^{[\ell]} \in  \mathbb{R}^{n_{\ell+1}\times n_{\ell}}, \bm{b}^{[\ell]}\in\mathbb{R}^{n_{\ell+1}}$ are the weights and biases of $\ell$-th hidden layer, respectively, $n_0=d$ and $n_{L+1}$ is the dimension of output, and $``\circ"$ stands for the elementary-wise operation. The function $\sigma(\cdot)$ is an element-wise activation function. We denote the output of a DNN by $\bm{y}(\bm{x};\bm{\theta})$ with $\bm{\theta}$ representing the parameter set of $\bm{W}^{[1]},\cdots \bm{W}^{[L]}, \bm{b}^{[1]},\cdots \bm{b}^{[L]}$.

Consequently, with the physical constraints, Bayes’ theorem under the
context of BPINN can be formulated as follows:
\begin{equation}
    P(\bm{\theta}|\mathcal{D}) = \frac{P(\mathcal{D}|\bm{\theta}) P(\bm{\theta})}{P(\mathcal{D})} \simeq P(\mathcal{D}|\bm{\theta}) P(\bm{\theta}),
\end{equation}
Then, the log-likelihood can be calculated as:
\begin{equation}\label{loglikelihood}
    \log P(\mathcal{D}|\bm{\theta}) = \log P(\mathcal{D}_f|\bm{\theta})+ \log 
 P(\mathcal{D}_g|\bm{\theta}),
\end{equation}
with 
\begin{equation}
    P(\mathcal{D}_f|\bm{\theta}) = \prod_{i=1}^{N_f} \frac{1}{\sqrt{2\pi\tau_f^2}} \exp\left(-\frac{\left\|\mathcal{N}_{\bm{\lambda}}\left[u_{NN}\left(\bm{x}_f^{(i)}; \bm{\theta}\right)\right] - \tilde{f}^{(i)}\right\|^2}{2\tau_f^2}\right)
\end{equation}
and
\begin{equation}
    P(\mathcal{D}_g|\bm{\theta}) = \prod_{j=1}^{N_g} \frac{1}{\sqrt{2\pi\tau_g^2}} \exp\left(-\frac{\left\|\mathcal{B}\left[u_{NN}\left(\bm{x}_g^{(j)};\bm{\theta}\right)\right] - \tilde{g}^{(j)}\right\|^2}{2\tau_g^2}\right).
\end{equation}
where $\|\cdot\|$ is the 2-norm for a given matrix or vector throughout the paper.

So far, we have specified the posterior distribution of BPINN. However, obtaining an explicit solution is often intractable. A traditional approach to approximate the posterior is to use sampling methods, such as Markov Chain Monte Carlo (MCMC). For instance, Hamiltonian Monte Carlo (HMC) (see Appendix \ref{App_HMC}) utilizes Hamiltonian dynamics to propose candidate parameter sets, followed by an acceptance-rejection mechanism to approximate the posterior density. Another widely used approach is variational inference (VI), which approximates the posterior by selecting a distribution $Q$ from a family of distributions $\mathcal{Q}$ and optimizing it to minimize the Kullback-Leibler (KL) divergence between $Q$ and the true posterior. This is typically achieved using a reparameterization trick, which allows efficient gradient-based optimization of the variational parameters. And the well-approximated posterior will be used to infer PDE's numerical solution.

\subsection{Failure of BPINN} \label{subsec:2.2 Failure}
HMC-driven BPINN method (called BPINN\_HMC) has shown its remarkable performance in solving general PDEs, as demonstrated in \cite{yang2021b}. However, the convergence of the HMC algorithm appears to be problematic when we try to extend BPINN to solve more complex PDE problems. The main problems are: 1) BPINN occasionally struggles to capture the primary signal in cases with highly oscillatory behavior, 2) sometimes HMC does not converge and provide us an available result.

To illustrate these two problems, we here introduce a concrete example in the context of multiscale elliptic PDEs. Let us consider the following one-dimensional elliptic equation with a homogeneous Dirichlet boundary in $\Omega = [0, 1]$:

\begin{equation}\label{multiscale_elliptic}
\begin{cases}
    \displaystyle -\frac{d}{dx}\left(A^\varepsilon(x)\frac{d}{dx} u^\varepsilon(x)\right) = f(x), \\
    u^\varepsilon(0) = u^\varepsilon(1) = 0,
\end{cases}
\end{equation}
in which 
\begin{equation}
A^\varepsilon(x) = \frac{1}{2 + \cos\left(2\pi x/\varepsilon\right)},
\end{equation}
with $\varepsilon > 0$ being a small constant and $f(x)=1$. Under these conditions, a unique solution is given by
\begin{equation}\label{DiffusionEq_1d_01_ueps}
u^\varepsilon(x) = x-x^2+\varepsilon\left(\frac{1}{4\pi}\sin\left(2\pi\frac{x}{\varepsilon}\right)-\frac{1}{2\pi}x\sin\left(2\pi\frac{x}{\varepsilon}\right)-\frac{\varepsilon}{4\pi^2}\cos\left(2\pi\frac{x}{\varepsilon}\right)+\frac{\varepsilon}{4\pi^2}\right).
\end{equation}

\begin{figure} [H]
	\centering  
	\vspace{-0.1cm}  
	\subfigure[$u^{\varepsilon}$ for $\varepsilon=0.5$] {
		\label{Test_BPINN_HMC2Eps0p1_Noise0p01}     
		\includegraphics[scale=0.425]{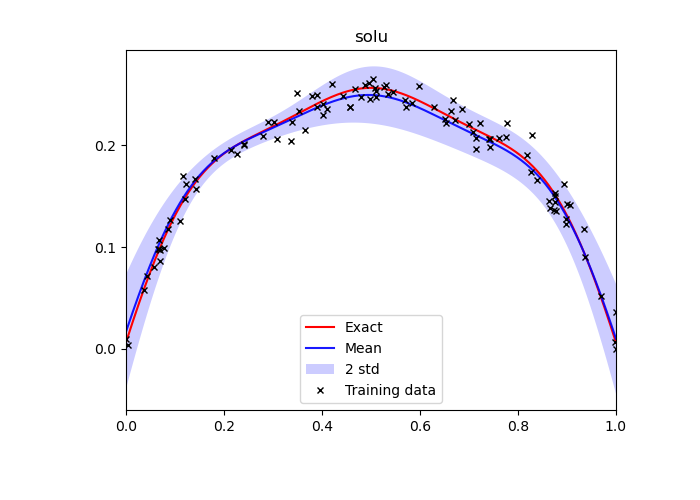}  
	}   
	\vspace{-0.1cm}  
	\subfigure[$u^{\varepsilon}$ for $\varepsilon=0.1$] {
		\label{Test_BPINN_HMC2Eps0p5_Noise0p01}     
		\includegraphics[scale=0.425]{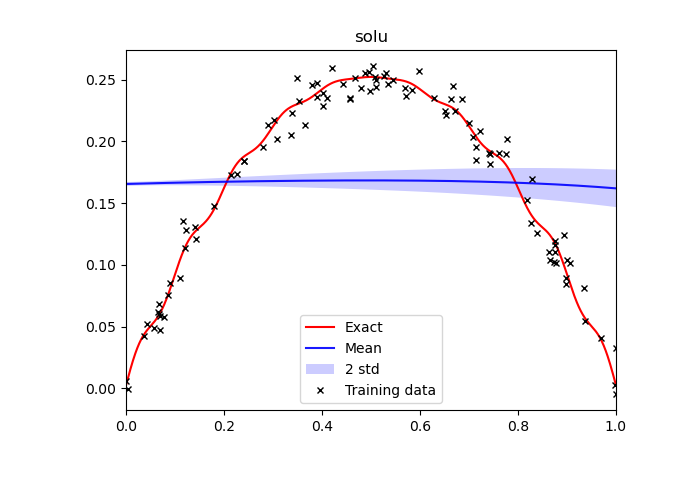}  
	}                   
    \caption{The predictions of BPINN\_HMC and exact solution for $\varepsilon=0.5$ and $\varepsilon=0.1$ when noise level is 0.01, respectively.} 
	\label{Test_BPINN_HMC2Noise0p01}         
\end{figure}
\begin{figure} [H]
	\centering   
	\vspace{-0.1cm}  
	\subfigure[$u^{\varepsilon}$ for $\varepsilon=0.5$ ] {
		\label{Test_BPINN_HMC2Eps0p1_Noise0p05}     
		\includegraphics[scale=0.425]{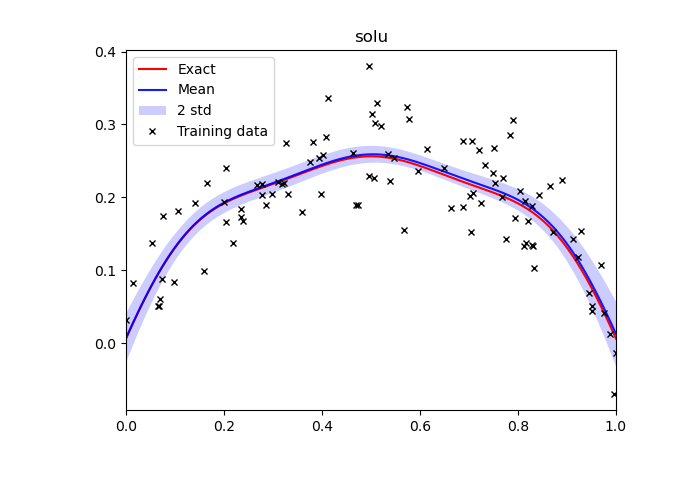}  
	}  
	\vspace{-0.1cm}  
	\subfigure[$u^{\varepsilon}$ for $\varepsilon=0.1$] {
		\label{Test_BPINN_HMC2Eps0p5_Noise0p05}     
		\includegraphics[scale=0.425]{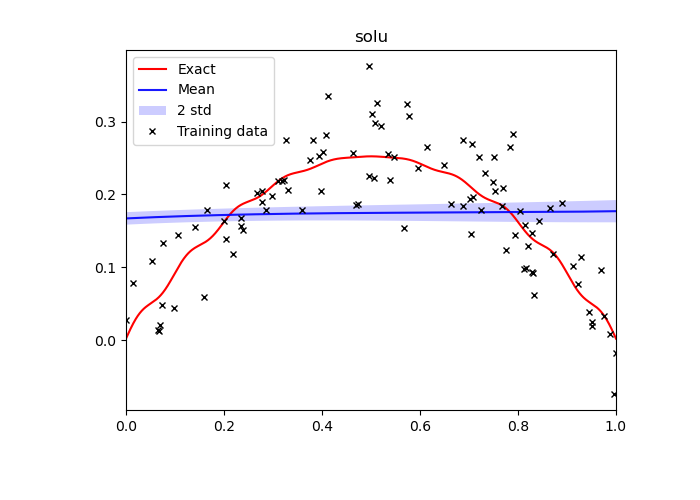}  
	}                   
	\caption{The predictions of BPINN\_HMC and exact solution for $\varepsilon=0.5$ when noise level is 0.05, respectively.} 
	\label{Test_BPINN_HMC2Noise0p05}         
\end{figure}
\begin{figure} [H]
	\centering   
	\vspace{-0.1cm}  
	\subfigure[$u^{\varepsilon}$ for $\varepsilon=0.5$ ] {
		\label{Test_BPINN_HMC2Eps0p1_Noise0p1}     
		\includegraphics[scale=0.425]{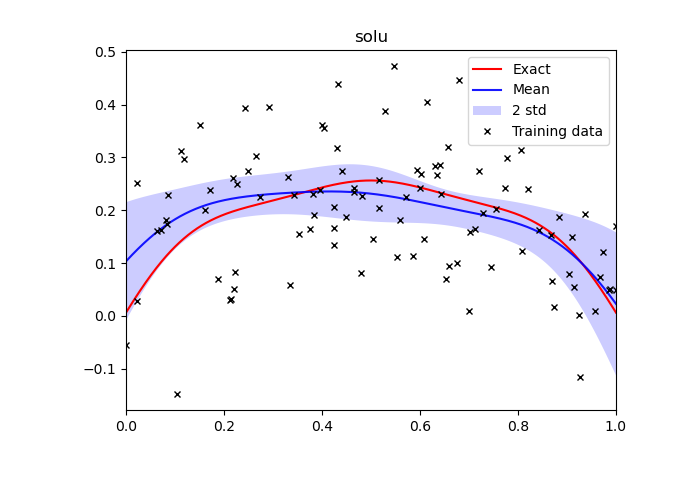}  
	}
	\vspace{-0.1cm}    
	\subfigure[$u^{\varepsilon}$ for $\varepsilon=0.1$] {
		\label{Test_BPINN_HMC2Eps0p5_Noise0p1}     
		\includegraphics[scale=0.425]{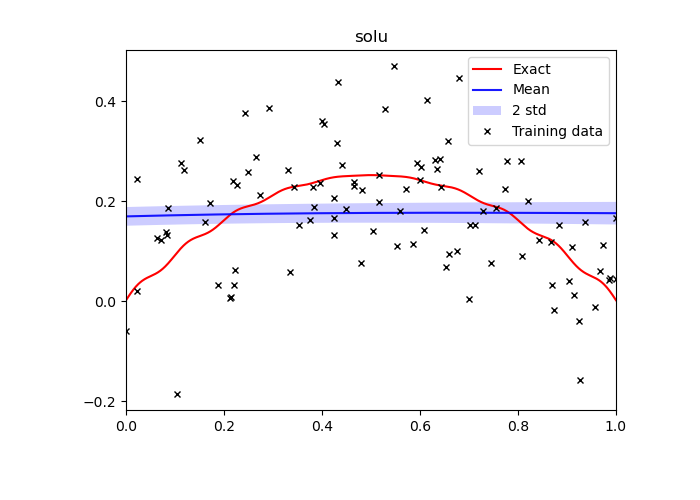}  
	}                   
	\caption{The predictions of BPINN\_HMC and exact solution for $\varepsilon=0.5$ when noise level 0.1, respectively.} 
	\label{Test_BPINN_HMC2Noise0p1}         
\end{figure}

We employ the above BPINN\_HMC method to solve the above problem when $\varepsilon=0.5$ and $\varepsilon=0.1$. For this study, we utilize 100 observations for system state and force side, respectively, in $\Omega$ to recover the solution of \eqref{multiscale_elliptic}. Meantime, 1000 equidistant points are sampled in $\Omega$ to evaluate this method. The solver of BPINN\_HMC
comprises two hidden layers and each layer contains 30 hidden units. The activation function used in all hidden layers is the sine function, while the output layer is linear. To estimate the internal parameters of the BPINN, we run the BPINN\_HMC method for 200 epochs and obtain the posterior distribution according to the results of the latter 150 epochs, the step number is 500 in each epoch and the step size used to update the internal parameters is fixed. Figures \ref{Test_BPINN_HMC2Noise0p01} -- \ref{Test_BPINN_HMC2Noise0p1} demonstrate the performance of BPINN-HMC for both smooth case ($\varepsilon=0.5$) and slight oscillation case ($\varepsilon=0.1$) when the noise level is 0.01, 0.05 and 0.1, respectively. It reveals that this classical method will fail to capture the exact solution for other oscillation cases. Herein and thereafter, The symbol `---' stands for the failure of HMC.

Furthermore, we perform BPINN\_HMC with various step sizes (0.05, 0.001, 0.0005, 0.0001, 0.00001) to evaluate its robustness when the noise level is 0.01, 0.05 and 0.1, respectively. The results, as shown in Tables \ref{NotRobust0p6} and \ref{NotRobust1}, indicate that the BPINN\_HMC method will fail to converge and encounter errors when the step size is large for different noise levels. This further underscores the limitations of the classical BPINN\_HMC approach in dealing with both smooth and oscillation scenarios. 

\vspace{-0.5cm}

\vspace{-0.5cm}

\vspace{-0.5cm}
\vspace{-0.25cm}

\begin{table}[H]
	\centering
	\caption{The performance of BPINN\_HMC with different fixed step size when $\varepsilon=0.5$.}
	\label{NotRobust0p6}
	\begin{tabular}{|l|c|c|c|c|c|c|c|c|c|c|}
		\hline
		Noise level &step size               &0.01     &0.005  &0.001    &0.0005    &0.0001       &0.00005       &0.00001     \\  \hline
\multirow{2}{*}{0.01}&Success/Failure  &Failure  &Failure&Failure  &Failure   &Failure      &Failure       &Success   \\  \cline{2-9}
        &REL of solution               &---      &---    &---      &---       & ---         &---           &0.0242  \\  \hline
\multirow{2}{*}{0.05}&Success/Failure  &Failure  &Failure&Failure  &Failure   &Success      &Success       &Success   \\   \cline{2-9}
        &REL of solution               &---      &---    &---      &---       &0.0220       &0.0177        &0.2567  \\  \hline
\multirow{2}{*}{0.1}&Success/Failure   &Failure  &Failure&Failure  &Failure   &Success      &Success       &Success   \\   \cline{2-9}
        &REL of solution               &---      &---    &---      &---       &0.0543       &0.1594        &0.3204  \\  \hline
	\end{tabular}
\end{table}

\begin{table}[H]
	\centering
	\caption{The performance of BPINN\_HMC with different fixed step size when $\varepsilon=0.1$.}
	\label{NotRobust1}
	\begin{tabular}{|l|c|c|c|c|c|c|c|c|c|c|}
		\hline
		Noisem level &step size               &0.01    &0.05      &0.001    &0.0005    &0.0001       &0.00005       &0.00001     \\  \hline
\multirow{2}{*}{0.01}&Success/Failure  &Failure &Failure   &Failure  &Failure   &Failure      &Failure       &Success   \\  \cline{2-9}
        &REL of solution               &---    &---        &---      &---       &---          &---           &0.3929 \\  \hline
\multirow{2}{*}{0.05}&Success/Failure  &Failure &Failure   &Failure  &Failure   &Failure      &Success       &Success   \\  \cline{2-9}
        &REL of solution               &---    &---        &---      &---       &---          &0.3958        &0.3979 \\  \hline
\multirow{2}{*}{0.1}&Success/Failure   &Failure &Failure   &Failure  &Failure   &Success      &Success       &Success   \\  \cline{2-9}
        &REL of solution               &---    &---        &---      &---       &0.4018       &0.4016        &0.3966  \\  \hline
	\end{tabular}
\end{table}

In addition, we perform the BPINN with different setups of the hyperparameters such as the number of sampled points and burn-in points as well as the network size, but we still cannot obtain a satisfactory result. Based on the above observation, it is necessary to probe into the misconvergence of HMC and develop new techniques to improve the accuracy and robustness of the BPINN method.

In this sensoria, the HMC may fail to converge in complex inference problems due to several challenges inherent to the posterior distribution and the algorithm's mechanics. First, HMC struggles with multimodal posterior distributions, where the parameter space contains multiple isolated regions of high probability. HMC generates samples by simulating the dynamics of a particle moving through the parameter space, and each HMC chain represents a sequence of such samples. In such cases, a single HMC chain can become trapped in one mode, unable to explore the entire posterior within a reasonable time \cite{zong2025randomized}.
Second, The high dimensionality of the parameter space exacerbates the difficulty of effective sampling, as high-dimensional posteriors often exhibit strong correlations and complex geometric structures that hinder efficient exploration by the Markov chains. This can lead to poor mixing and slow transitions between regions of high probability. As shown in Figure \ref{Results2Multiscale2D_Eps0p5}, the point-wise error for HMC displays significant variability, with noticeable error peaks indicating regions where the posterior distribution is not well captured. These limitations highlight the difficulty of using HMC in high-dimensional settings, especially when the posterior geometry complicates sampling and convergence.
Additionally, high curvature in certain areas of the posterior can cause numerical instability in HMC's leapfrog integrator, further impeding convergence. For example, as shown in Appendix \ref{App_HMC}, the momentum is updated using the $param\_grad$ function, where momentums of the parameters are calculated. Poorly scaled gradients or an imbalanced parameter space often lead to excessively large momentum values, which propagate through the leapfrog updates, causing the likelihood of the parameters to explode to infinity. This instability is exacerbated in high-dimensional or complex, non-linear problems, where the posterior distribution exhibits sharp gradients or strong correlations. These challenges emphasize the importance of careful tuning of HMC hyperparameters, such as step size and trajectory length, which become increasingly sensitive under such conditions. As a result, HMC may fail to provide a representative sample of the posterior, leading to biased estimates and unreliable inference.





\section{Methodology}\label{sec:03}

\subsection{SGD Reframed HMC method} \label{subsec:3.2 SGD}
In this section, to avoid the convergence failure of classic HMC, we would like to use SGD to reframe the existing method. Unlike HMC, where the entire posterior distribution of the parameters is modeled, we aim to find the set of parameters that are most ``likely'', as this procedure guarantees a solution. Notice that, in each step of HMC, a likelihood is calculated for a set of parameters and is used to determine whether this step will be accepted. Finally, all accepted steps form the empirical posterior distribution. The algorithm of HMC is as in Appendix \ref{App_HMC}. 

Through the HMC algorithm Appendix \ref{App_HMC}, we could see that each set of parameters is considered as a sample from the numerator of the posterior distribution, also referred to as the kernel function of BPINN. We would like to define the most ``likely" set of parameters as the set that maximized the kernel function. Thus, by finding the most ``likely" set of parameters, we mean seeking to find the maximum of the true distribution by viewing the inverse of the likelihood as the loss. The larger the likelihood, the smaller the loss. This turns the problem into an optimization problem, and we use typical stochastic gradient descent (SGD) to find the optimum.

\begin{algorithm}[H]
\caption{Adam optimization method} \label{alg:adam}
\begin{algorithmic}[1]
\REQUIRE  Stepsize $\alpha$ and Exponential decay rates for the moment estimates $\beta_1, \beta_2\in(0,1]$
\REQUIRE  Stochastic objective function $f(\bm{\theta})$ with parameters $\bm{\theta}$
\renewcommand{\algorithmicensure}{\textbf{Input:}}
\ENSURE Initialize the parameter vector $\bm{\theta}_0\neq \bm{0}$, the $1^{st}$ moment vector $\bm{m}_0 = \bm{0}$, the $2^{nd}$ moment vector $\bm{v}_0= \bm{0}$ and timestep $t=0$
\WHILE{$\theta_t$ not converged} 
 \STATE t = t+1
 \STATE $\bm{g}_t = \nabla_{\bm{\theta}} f_t(\bm{\theta}_{t-1})$
 \STATE $\bm{m}_t=\beta_1\cdot \bm{m}_{t-1} + (1-\beta_1)\cdot \bm{g}_t$
 \STATE $\bm{v}_t=\beta_2\cdot \bm{v}_{t-1} + (1-\beta_2)\cdot \bm{g}_t\odot \bm{g}_t$
 \STATE $\widehat{\bm{m}}_t = \bm{m}_t/(1-\beta_1^t)$
 \STATE $\widehat{\bm{v}}_t = \bm{v}_t/(1-\beta_2^t)$
 \STATE $\bm{\theta}_t=\bm{\theta}_{t-1}-\alpha\cdot\widehat{\bm{m}}_t/(\sqrt{\widehat{\bm{v}}_t}+\varepsilon)$
\ENDWHILE
\RETURN $\bm{\theta}_t$
\end{algorithmic}
\end{algorithm}

\subsection{MBPINN} \label{subsec:3.1 MBPINN}
In this section, the unified architecture of MBPINN is proposed to estimate the parameters and recover the solution of PDEs according to the given observation data with unexpected noise by embracing the vanilla BPINN method with Fourier-induced MscaleDNN.

In terms of the BPINN method, its solver is configured as a general DNN. As we are aware, a general DNN model is capable of providing a satisfactory solution for low-complexity problems but will encounter troublesome difficulty in solving complex problems such as multi-scale PDEs. From the perspective of spectral bias or frequency preference, the DNN is typically efficient for fitting objective functions with low-frequency modes but inefficient for high-frequency functions. Therefore, the PINN method have the same phenomenon, it will have an adverse impact for the performance of BPINN. Neural Tangent Kernel (NTK) theory has been used to model and understand the behavior of DNN\cite{wang2020eigenvector}.
  
Recently, a MscaleDNN architecture has been proposed based on the intrinsic property of DNN to mitigate the pathology of DNN by converting original data to a low-frequency space \cite{liu2020multi,wang2020multiscale,wang2024practical,li2023subspace}. According to the NTK theory, the authors in \cite{wang2020eigenvector} developed a random Fourier feature mapping (FFM) induced Physics-informed neural networks  to enhance the robustness and accuracy of conventional PINN for solving multi-scale PDEs, this new framework is another type of MscaleDNN. The FFM  is expressed as follows  
\begin{equation*}
\zeta(\bm{x}) = \Bigg{[}\begin{matrix}
\cos(\pi \Lambda \bm{x}) \\ \sin(\pi \Lambda \bm{x})
\end{matrix}\Bigg{]}
\end{equation*}
where $\Lambda$ is a transmitted matrix that is consistent with the dimension of input data and the number of neural units for the first hidden layer in DNN, and its elements are sampled from an isotropic Gaussian distribution $\mathcal{N}(0, \xi)$ with $\xi>0$ is a user-specified hyper-parameter. Numerical simulations demonstrated the heuristic sinusoidal mapping of input will enable DNN to efficiently fit higher frequency functions. Moreover, the authors in \cite{li2023deep} peoposed a modified FFM that keeps the original independent variables as the input to balance the ability of DNN and PINN for dealing with low frequencies and high frequency components. In this paper, we modify this strategy by utilizing a linear transformation for the inputs data, it is 
\begin{equation*}
\zeta(\bm{x}) = \Bigg{[}\begin{matrix}
\textrm{Lin}(\bm{x})\\
\cos(\pi \Lambda \bm{x}) \\ \sin(\pi \Lambda \bm{x})
\end{matrix}\Bigg{]}
\end{equation*}
where $Lin(\bm{x}) = \bm{W}_{lin}\bm{x}+\bm{b}_{lin}$ with  $\bm{W}_{lin}$ bing the weight matrix and $\bm{b}_{lin}$ bing the bias vector. 

Hence, we can improve the capacity of BPINN by embracing the multi-scale DNN with the above modified FFM , called it as MBPINN method. A schematic diagram of MscaleDNN with multiple modified FFM pipelines is described as follows.
\begin{equation}\label{MFFM_DNN}
\begin{aligned}
 \hat{\bm{x}}&= \bm{\Lambda}_n\bm{x}, \quad n=1,2, \ldots, N,\\
 \bm{\zeta}_{n}(\bm{x}) & =\left[\begin{matrix}
 	\textrm{Lin}(\bm{x})\\
    \cos \left( \pi\hat{\bm{x}}\right)\\
    \sin \left(\pi\hat{\bm{x}}\right)
 \end{matrix}\right], \quad n=1,2, \ldots, N, \\
 \boldsymbol{F}_n(\bm{x}) & =\mathcal{F C N}\left(\bm{\zeta}_{n}(\bm{x})\right), \quad n=1,2, \ldots, N, \\
 \boldsymbol{NN}(\bm{x}) & =\boldsymbol{W}_O \cdot\left[\boldsymbol{F}_{1}, \boldsymbol{F}_{2}, \cdots, \boldsymbol{F}_{N}\right]+\boldsymbol{b}_O,
\end{aligned}
\end{equation}
in which, $\mathcal{FCN}$ stands for a fully connected neural network.  $\boldsymbol{W}_O$ and $\boldsymbol{b}_O$ represent the weights and bias for the output layer of MscaleDNN, respectively. 

\begin{figure}
	\centering
		\includegraphics[scale=0.325]{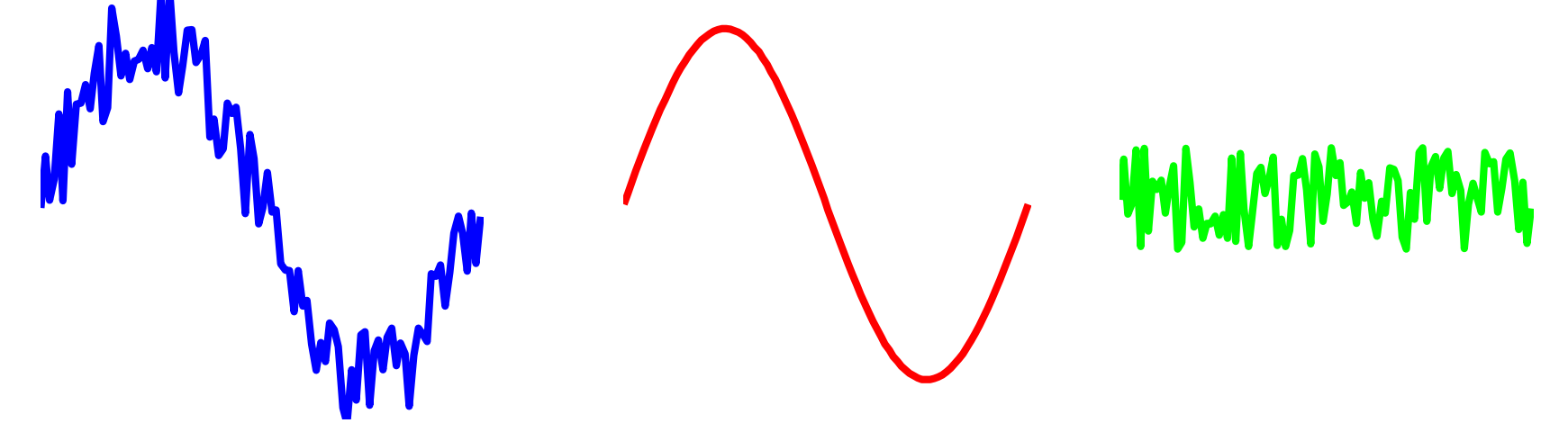}
	\caption{The decomposition of multi-scale function. Left: Original function, Middle: Coarse part, Right: Oscillation part}
	\label{fig2multiscale_solu}
\end{figure}

For a given multi-scale PDE, the solution generally has the following coarse/fine decomposition, $u=u_c+u_f$, in which $u_c$ contains the smooth part and $u_f$ contains the fine details of the multi-scale solution $u$, respectively. Please see figure \ref{fig2multiscale_solu}.
Naturally, the force term $f$ and the boundary constraint $g$ may also be oscillating and could be decomposed as $f = f_c+f_f$ and $g = g_c+g_f$, respectively. Analogously, we have the collection set for governed equation and boundary constraint, i.e., $\mathcal{D}_f = \left\{\left(x_f^{(i)}, \tilde{f}^{(i)}\right)\right\}_{i=1}^{N_f}$ and $\mathcal{D}_g = \left\{\left(\bm{x}_g^{(j)}, \tilde{g}^{(j)}\right)\right\}_{j=1}^{N_g}$.  These observation are i.i.d Gaussian random variables and thay have the following formulations
\begin{align}
    \tilde{f}^{(i)} &= f_c(x_f^{(i)}) + f_f(x_f^{(i)})+ \varepsilon_f^{(i)}, \quad x_f^{(i)} \in \mathcal{D}_f~ \textup{and} ~i = 1, 2, \ldots, N_f, \\
    \tilde{g}^{(j)} &= g_c(x_g^{(j)}) + g_f(x_g^{(j)}) + \varepsilon_g^{(j)}, \quad x_g^{(j)} \in \mathcal{D}_g  ~ \textup{and} ~ j = 1, 2, \ldots, N_g.
\end{align}
As well as, the $\bm{\varepsilon}_f$ and $\bm{\varepsilon}_g$ are  same as the aforementioned definitions in section \ref{subsec:BPINN}. 


Within the framework of MBPINN, the log-likelihood can be calculated as follows:

\begin{equation}
    \log P(\mathcal{D}|\bm{\theta}) = \log P(\mathcal{D}_f|\bm{\theta})+ \log 
 P(\mathcal{D}_g|\bm{\theta}),
\end{equation}
with 
\begin{equation}
    P(\mathcal{D}_f|\bm{\theta}) = \prod_{i=1}^{N_f} \frac{1}{\sqrt{2\pi\tau_f^2}} \exp\left(-\frac{\left\|\mathcal{N}_{\bm{\lambda}}\left[u_{NN}\left(\bm{x}_f^{(i)}; \bm{\theta}\right)\right] - \tilde{f}^{(i)}\right\|^2}{2\tau_f^2}\right)
\end{equation}
and
\begin{equation}
    P(\mathcal{D}_g|\bm{\theta}) = \prod_{j=1}^{N_g} \frac{1}{\sqrt{2\pi\tau_g^2}} \exp\left(-\frac{\left\|\mathcal{B}\left[u_{NN}\left(\bm{x}_g^{(j)};\bm{\theta}\right)\right] - \tilde{g}^{(j)}\right\|^2}{2\tau_g^2}\right).
\end{equation}

\begin{figure}[H]
	\centering
	\includegraphics[scale=0.425]{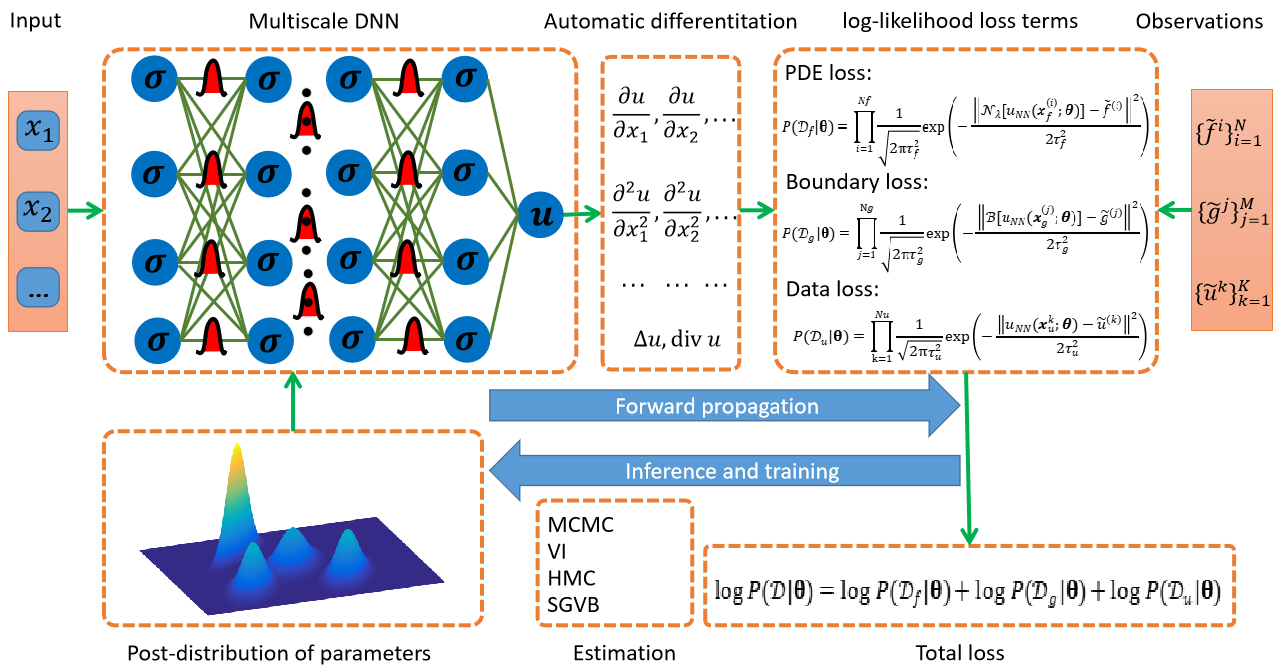}
	\caption{The semantic diagram for MBPINN to solve multi-scale PDEs with noise}
	\label{fig2mscalednn}
\end{figure}

If some additional observed data are available inside the interested domain, it is $\mathcal{D}_u = \left\{\left(x_u^{(k)}, \tilde{u}^{(k)}\right)\right\}_{k=1}^{N_u}$ with $ \tilde{u}^{(k)} = u_c(x_u^{(k)}) + u_f(x_u^{(k)}) + \varepsilon_u^{(k)}, ~k = 1, 2, \ldots, N_u$, then a log-likelihood term indicating the mismatch between the predictions produced by MBPINN and the observations can be taken into account
\begin{equation}
    P(\mathcal{D}_u|\bm{\theta}) = \prod_{k=1}^{N_u} \frac{1}{\sqrt{2\pi\tau_u^2}} \exp\left(-\frac{\left\|u_{NN}\left(x_u^{(k)}; \bm{\theta}\right) - \tilde{u}^{(k)}\right\|^2}{2\tau_u^2}\right)
\end{equation}


where $\tau_f$ is the standard deviations for mean-zero Gaussian noise $\bm{\varepsilon}_u$. At the end, we conclude the section with the schematic of the MBPINN architecture in Figure \ref{fig2mscalednn}.

\section{Experiment}\label{sec:04}
The goal of our experiments is to show that our MBPINN with SGD reframed HMC is indeed capable of approximating the analytical solution for given general and complex PDE based on observations with various noise levels. In addition, the BPINN method with a DNN model being its solver is introduced to serve as the baseline. Six types of compared methods are described in the following:
\begin{itemize}
    \item \emph{BPINN\_HMC}: Its solver is a normal DNN model with the Sine function being the activation function for all hidden layers, while the output layer is linear. The approach of estimating its parameters is classical HMC \cite{yang2021b}.
    \item \emph{BPINN\_SGD}: Its solver is a normal DNN model with the Sine function being the activation function for all hidden layers, while the output layer is linear. The approach of estimating its parameters is SGD (Adam).
    \item \emph{FF\_MBPINN\_HMC}: Its solver is an MscaleDNN model with an FFM pipeline, the activation function for the remainder hidden layers is chosen as Sine, while the output layer is linear.  The approach to estimating its parameters is classical HMC.
    \item \emph{FF\_MBPINN\_SGD}: Its solver is a MscaleDNN model with an FFM pipeline, the activation function for the remainder hidden layers is chosen as Sine, while the output layer is linear. The approach to estimating its parameters is SGD (Adam).
    \item \emph{2FF\_MBPINN\_HMC}: Its solver is an MscaleDNN model with two FFM pipelines, the activation function for the remainder hidden layers is chosen as Sine, while the output layer is linear. The approach to estimating its parameters is classical HMC.
    \item \emph{2FF\_MBPINN\_SGD}: Its solver is a MscaleDNN model with two FFM pipelines, the activation function for the remainder hidden layers is chosen as Sine, while the output layer is linear. The approach to estimating its parameters is SGD (Adam).
\end{itemize}
For comparison, we run HMC algorithm for 200 trajectors, then record the last 150 trajectors and obtain the posterior distribution. In each trajector, the step number to update network's parameters is set as 500 and the step size is fixed. In addition, we perform the SGD (Adam) used in the above methods with a fixed learning rate for 20000 epochs and obtain the most 'likely' parameters of the FF\_MBPINN\_SGD and 2FF\_MBPINN\_SGD methods according to the final epoch's results. In our test, the step size or learning rate are set as 0.01, 0.005, 0.001, 0.0005, 0.0001, 0.00005 and 0.00001, respectively.

As for the observations, we generate them based on random sampling strategy (For example, Latin Hypercube Sampling (LHS)) in the interest domain. In which, LHS is a statistical method that provides a more randomized distribution of points, particularly useful for high-dimensional spaces, while ensuring boundary points are included and corresponding values are computed with noise. Finally, we add noise based on different noise levels. Noise level is defined by the constant times of a standard normal distribution. For example, the noise level of 0.1 means $0.1 * \mathcal{N}(0, 1)$. In addition, we utilize the signal-to-noise ration (SNR) to describe the influence of noise for solution, coefficient term and force size, respectively. It is
\begin{equation*}
\label{snr}
   SNR = 10\log\left(\frac{Ps}{Pn}\right)
\end{equation*}
where $Ps$ and $Pn$ stand for the signal power  and noise power, respectively. 

    

To quantitatively measure the performance of our model, we compute the point-wise Absolute Error (ABSE) and the Relative Error (REL) between the mean-predicted and the exact solutions. The ABSE is given by
    \[
    \text{ABSE} = | \hat{u}(x_i') - u^*(x_i') |, ~~i = 1, 2, \dots, N'
    \]
    and the REL is calculated as follows:
    \[
    \text{REL} = \sqrt{\frac{\sum_{i=1}^{N'} |\hat{u}(x_i') - u^*(x_i')|^2}{\sum_{i=1}^{N'} |u^*(x_i')|^2}},
    \]
    where $\hat{u}(x_i')$ denotes the mean-predicted solution at point $x_i'$, $u^*(x_i')$ is the exact solution, and $N'$ is the total number of evaluation points. 

\subsection{Performance of MBPINN for Solving 1-dimensional PDEs}

\begin{figure} [H]
	\centering 
	 \vspace{-0.2cm}  
	\subfigure[BPINN\_HMC] {
		\label{BPINNHMC2Multiscale1D_Noise0p01}     
		\includegraphics[scale=0.325]{Test_BPINN_HMC2Solu0p01_Eps0p1.png} 
	}   
	\vspace{-0.2cm}    
	\subfigure[BPINN\_SGD] {
		\label{BPINNSGD2Multiscale1D_Noise0p01}     
		\includegraphics[scale=0.325]{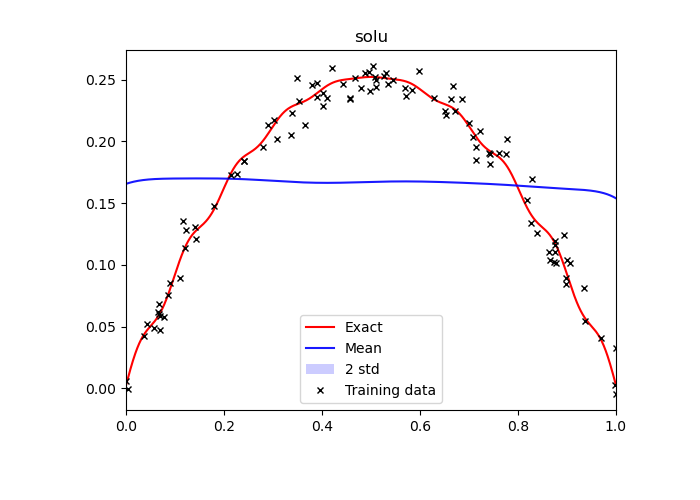}  
	}   
\vspace{-0.2cm}    
	\subfigure[FF\_MBPINN\_SGD] {	 
		\label{MBPINNSGD2Multiscale1D_Noise0p01}     
		\includegraphics[scale=0.325]{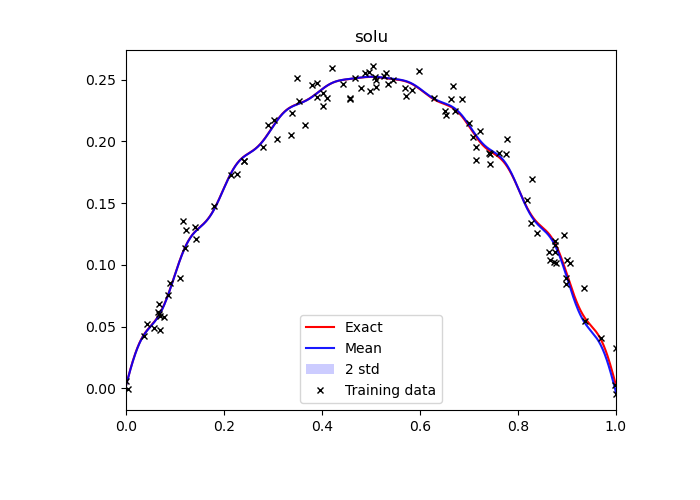}  
	}  
\vspace{-0.2cm} 
	\subfigure[2FF\_MBPINN\_SGD] {
		\label{MBPINNHMC2Multiscale1D_Noise0p01}     
		\includegraphics[scale=0.325]{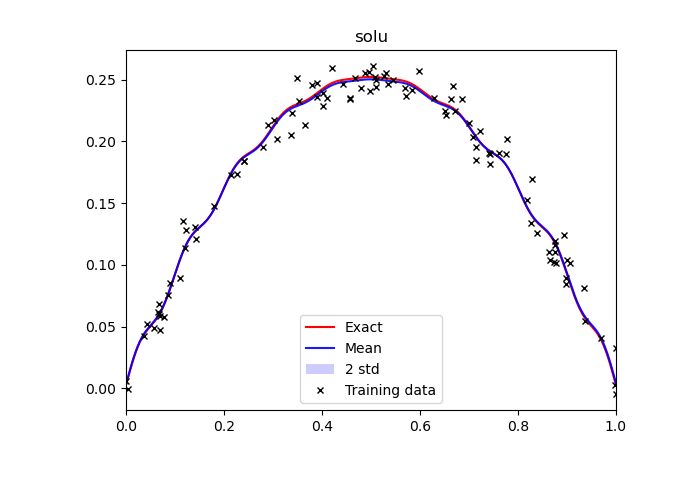}  
	}  
	\caption{The exact and predicted solutions for Example \ref{multiscale_elliptic1d:PDE2} when the noise level is 0.01 .} 
	\label{HMC_SGD_MBPINN2Multiscale1D_Noise0p01}         
\end{figure}

\begin{figure} [H]
	\centering 
	\vspace{-0.2cm}    
	\subfigure[BPINN\_HMC] {
		\label{BPINNHMC2Multiscale1D_Noise0p05}     
		\includegraphics[scale=0.325]{Test_BPINN_HMC2Solu0p05_Eps0p1.png}  
	}   
	\vspace{-0.2cm} 
	\subfigure[BPINN\_SGD] {
	\label{BPINNSGD2Multiscale1D_Noise0p05}     
		\includegraphics[scale=0.325]{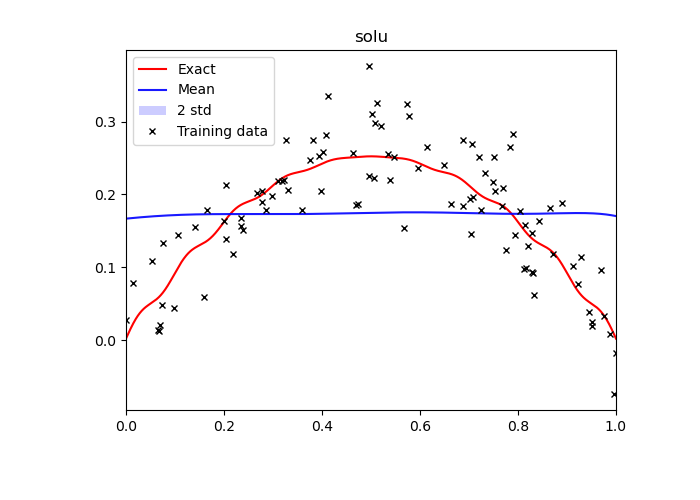}  
	} 
	\vspace{-0.2cm}        
	\subfigure[FF\_MBPINN\_SGD] {
		\label{MBPINNSGD2Multiscale1D_Noise0p05}     
		\includegraphics[scale=0.325]{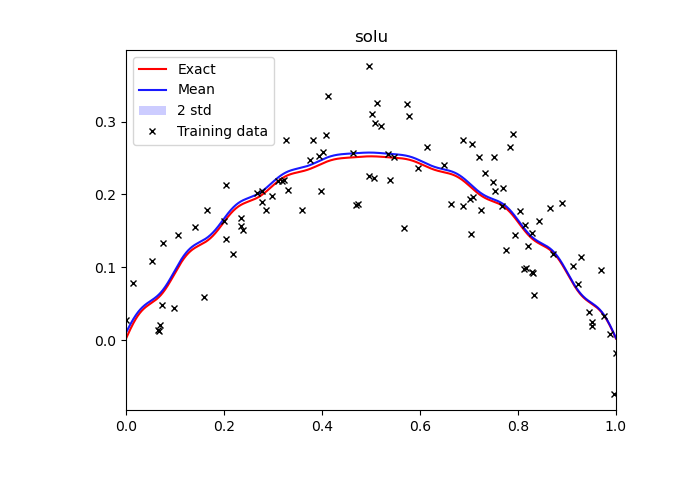}  
	}   
	\vspace{-0.2cm} 
	\subfigure[2FF\_MBPINN\_SGD] {
		\label{MBPINNHMC2Multiscale1D_Noise0p05}     
		\includegraphics[scale=0.325]{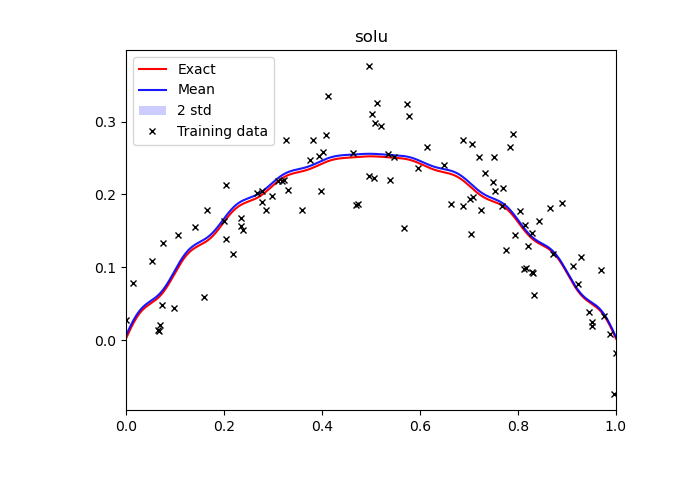}  
	}  
	\caption{The exact and predicted solutions  for Example \ref{multiscale_elliptic1d:PDE2} when the noise level is 0.05.} 
	\label{HMC_SGD_MBPINN2Multiscale1D_Noise0p05}         
\end{figure}

\begin{figure} [H]
	\centering
	\vspace{-0.2cm}     
	\subfigure[BPINN\_HMC] {
		\label{BPINNHMC2Multiscale1D_Noise0p1}     
		\includegraphics[scale=0.325]{Test_BPINN_HMC2Solu0p1_Eps0p1.png}  
	} 
	\vspace{-0.2cm}     
	\subfigure[BPINN\_SGD] {
		\label{BPINNSGD2Multiscale1D_Noise0p1}     
		\includegraphics[scale=0.325]{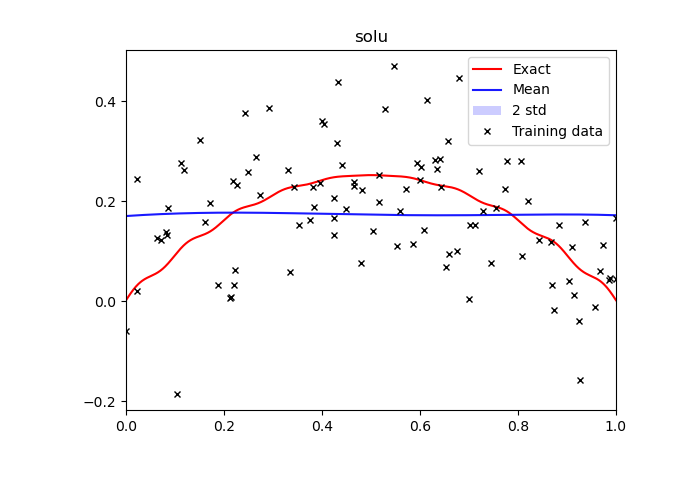}  
	} 
	\vspace{-0.2cm}          
	\subfigure[FF\_MBPINN\_SGD] {
		\label{MBPINNSGD2Multiscale1D_Noise0p1}     
		\includegraphics[scale=0.325]{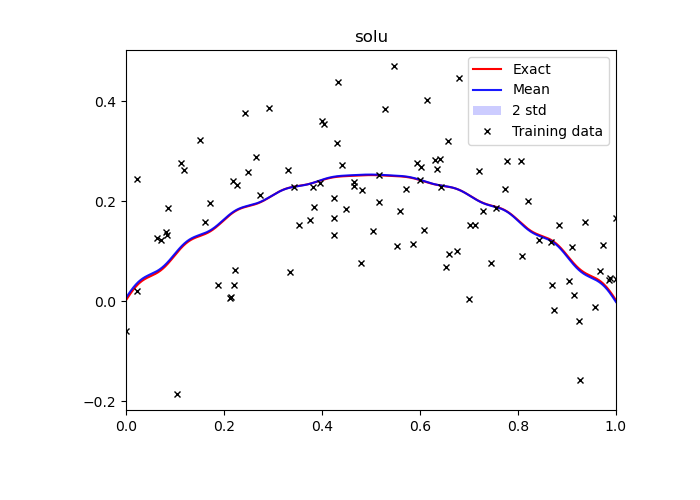}  
	}
	\vspace{-0.2cm}    
	\subfigure[2FF\_MBPINN\_SGD] {
		\label{MBPINNHMC2Multiscale1D_Noise0p1}     
		\includegraphics[scale=0.325]{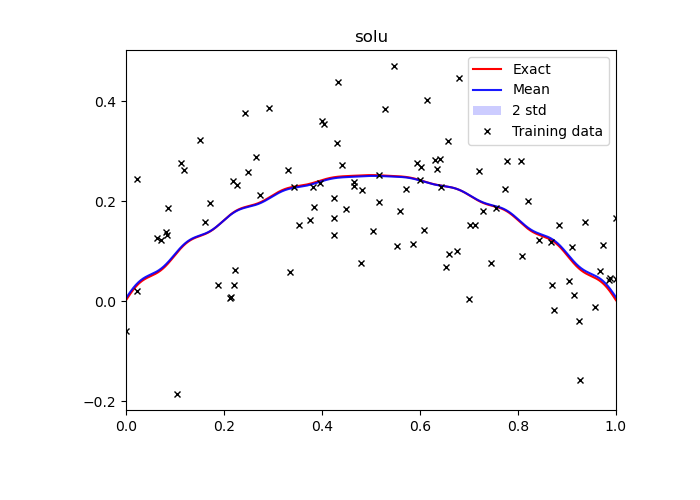}  
	}  
	\caption{The exact and predicted solutions for Example \ref{multiscale_elliptic1d:PDE2} when the noise level is 0.1.} 
	\label{HMC_SGD_MBPINN2Multiscale1D_Noise0p1}   
	\vspace{-0.2cm}        
\end{figure}

\begin{example}\label{multiscale_elliptic1d:PDE2}
	In this example, the problem setups are same as the section \ref{subsec:2.2 Failure} and we solve this  two-scale elliptic problem when $\varepsilon=0.1$ by employing the aforementioned six methods that their solvers have two hidden layers and each layer has 30 units. In our test, we only disturb the observation of the solution with a mean-zero Gaussian distributed noise on 98 inner random sampled points and 2 boundary points. As for the observations of force term, 100 observations located randomly in $\Omega$ are disturbed with mean-zero Gaussian distributed noise. To approximate the solution, the standard deviation for $\Lambda$ of MscaleDNN in FF\_MBPINN is set as 5,  the standard deviations for $\Lambda_1$ and $\Lambda_2$ of MscaleDNN in 2FF\_MBPINN are set as 1 and 5, respectively. Moreover, in order to showcase the performance of our proposed method, the SNRs for solution and force side under various noise levels are listed in Table \ref{snr2multiscale_elliptic1d}. We list and depict the related experiment results in Tables \ref{Table:solu2multiscale0p01} -- \ref{Table:solu2multiscale0p1} and Figures \ref{HMC_SGD_MBPINN2Multiscale1D_Noise0p01} -- \ref{HMC_SGD_MBPINN2Multiscale1D_Noise0p1}, respectively.
\end{example}
\vspace{-0.4cm}
\begin{table}[!ht]
	\centering
	\caption{SNR of solution and force side under various noise level for  Example \ref{multiscale_elliptic1d:PDE2}.}
	\vspace{-0.1cm}
	\label{snr2multiscale_elliptic1d}
	\begin{tabular}{|l|c|c|c|}
		\hline
		Noise level      &$0.01$           &$0.05$        &$0.1$         \\  \hline
		solution	     &24.527           &12.181       &5.008  \\ \hline
		force side       &39.537           &27.193      &20.811  \\  \hline
	\end{tabular}
\vspace{-0.4cm}
\end{table}

\begin{table}[H]
	\centering
	\caption{REL of different method to solve Example \ref{multiscale_elliptic1d:PDE2} when the noise level is 0.01.}
	\vspace{-0.25cm}
	\begin{tabular}{|l|ccccccc|}
		\hline
		Initial step size(lr)& 0.01 & 0.005 & 0.001 & 0.0005 & 0.0001 & 0.00005 & 0.00001\\ \hline
		BPINN\_HMC       & ---    & ---     & ---   & ---    &---     &---      &0.3929 \\
		FF\_MBPINN\_HMC  & ---    & ---     & ---   & ---    &---     &---      &--- \\
		2FF\_MBPINN\_HMC & ---    & ---     & ---   & ---    &---     &---      &--- \\ \hline
		BPINN\_SGD       &0.3931  &0.3957   &0.3854 &0.3873  &0.3953  &0.3965   &0.3943 \\
		FF\_MBPINN\_SGD  & 0.2132 &0.0892   &0.0083 &0.0198  &0.0174  &0.0195   &0.3014\\
		2FF\_MBPINN\_SGD & 0.3125 &0.0062   &0.0131 &0.0076  & 0.0202 &0.0419   &0.1705 \\ \hline
	\end{tabular}
\vspace{-0.3cm}
	\label{Table:solu2multiscale0p01}
\end{table}

\begin{table}[H]
	\centering
	\caption{REL of the different method to solve Example \ref{multiscale_elliptic1d:PDE2} when the noise level is 0.05.}
	\vspace{-0.25cm}
	\begin{tabular}{|l|ccccccc|}
		\hline
		Initial step size(lr)& 0.01  & 0.005 & 0.001 & 0.0005  & 0.0001 & 0.00005 & 0.00001\\ \hline
		BPINN\_HMC           & ---   & ---   & ---   & ---     &---     &0.3958   &0.3979\\
		FF\_MBPINN\_HMC      & ---   & ---   & ---   & ---     &---     &---      &--- \\
		2FF\_MBPINN\_HMC     & ---   & ---   & ---   & ---     &---     &---      &--- \\ \hline
		BPINN\_SGD           &0.3964 &0.3962 &0.3961 &0.3987   &0.3991  &0.3995   &0.3981 \\
		FF\_MBPINN\_SGD      &0.0876 &0.0663 &0.0323 &0.0223   &0.0464  &0.0747   &0.6851 \\
		2FF\_MBPINN\_SGD     &0.0174 &0.0405 &0.0201 &0.0216   &0.0257  &0.0342   &0.0911\\ \hline
	\end{tabular}
\vspace{-0.3cm}
	\label{Table:solu2multiscale0p05}
\end{table}

\begin{table}[H]
	\centering
	\caption{REL of the different method to solve Example \ref{multiscale_elliptic1d:PDE2} when the noise level is 0.1.}
	\vspace{-0.25cm}
	\begin{tabular}{|l|ccccccc|}
		\hline
		Initial step size(lr)& 0.01   & 0.005 & 0.001 & 0.0005 & 0.0001& 0.00005 & 0.00001\\ \hline
		BPINN\_HMC           & ---    & ---   & ---   & ---    &0.4018 &0.4016   &0.3966    \\
		FF\_MBPINN\_HMC      & ---    & ---   & ---   & ---    &---    &---      &--- \\
		2FF\_MBPINN\_HMC     & ---    & ---   & ---   & ---    &---    &---      &--- \\ \hline
		BPINN\_SGD           &0.4025  &0.4016 & 0.4018&0.4019  &0.4017 &0.4016   &0.3986\\
		FF\_MBPINN\_SGD      &0.0980  &0.1021 &0.0431 &0.0271  &0.0225 &0.0107   &0.4926\\
		2FF\_MBPINN\_SGD     &0.1073  &0.0125 &0.0409 &0.0505  &0.0393 &0.0288   &0.0618\\ \hline
	\end{tabular}
\vspace{-0.3cm}
	\label{Table:solu2multiscale0p1}
\end{table}

\vspace{-0.2cm}

\vspace{-0.25cm}

\vspace{-0.25cm}

The results in Tables \ref{Table:solu2multiscale0p01} -- \ref{Table:solu2multiscale0p1} and Figures \ref{HMC_SGD_MBPINN2Multiscale1D_Noise0p01} -- \ref{HMC_SGD_MBPINN2Multiscale1D_Noise0p1} demonstrate that the FF\_MBPINN\_SGD and 2FF\_MBPINN\_SGD methods  exceed the performance of the classical BPINN\_HMC and BPINN\_SGD methods when the HMC driven BPINN is worked. In addition, the performance of FF\_MBPINN\_SGD is competed against that of 2FF\_MBPINN\_SGD.  It is important to note that when the FFM is introduced into the classical BPINN framework, the convergence issues associated with HMC become even more pronounced because of the FFM layer enlarges the effects of stepsize for HMC's lepfrog.

\emph{Influence of the learning rate:} We study the influence of decayed step size or learning rate for our proposed methods to solve the above two-scale elliptic problem when the $\varepsilon=0.1$ and the noise level is 0.05. In the test, the initial value of step size or learning are set as 0.01, 0.005, 0.001, 0.0005, 0.0001, 0.00005 and 0.00001, respectively. In which, the learning rate will be decayed by 5\% for every 100 training epochs and the step size will decay 5$\%$ for every 20 steps in our traing processes. Their other setups are identical to the above experiments. Based on the results in Table \ref{Table2learning} and the results in Table \ref{Table:solu2multiscale0p05} , the decayed step size or learning rate is inferior to the fixed cases for the our proposed methods in solving multi-scale problems.

\vspace{-0.2cm}
\begin{table}[H]
	\centering
	\caption{REL of the different learning rates for the above methods to solve Example \ref{multiscale_elliptic1d:PDE2}.}
	\vspace{-0.2cm}
	\begin{tabular}{|l|ccccccc|}
		\hline
		Initial step size(lr)& 0.01  & 0.005 & 0.001 & 0.0005  & 0.0001 & 0.00005 & 0.00001\\ \hline
		BPINN\_HMC           & ---   & ---   & ---   & ---     &---     &0.3998   &0.3981\\
		FF\_MBPINN\_HMC      & ---   & ---   & ---   & ---     &---     &---      &--- \\
		2FF\_MBPINN\_HMC     & ---   & ---   & ---   & ---     &---     &---      &--- \\ \hline
		BPINN\_SGD           &0.3989 &0.3993 &0.3994 &0.3991   &0.3981  &0.3981   &0.3979\\
		FF\_MBPINN\_SGD      &0.1271 &0.0747 &0.0892 &1.0366   &6.1769  &7.0061   &6.3309 \\
		2FF\_MBPINN\_SGD     &0.0202 &0.0361 &0.1851 &0.2523   &0.4349  &1.5280   &2.7352\\ \hline
	\end{tabular}
	\label{Table2learning}
	\vspace{-0.2cm}  
\end{table}

\emph{Influence of the hidden units:}  We study the influence of hidden units including width and depth for our proposed methods to solve the aforementioned two-scale elliptic problem when the $\varepsilon=0.1$ and the noise level is 0.05. In the test, the hidden units are set as $(20,20)$, $(40,40)$,$(50,50)$, $(20, 20,20)$, $(30,30,30)$ and $(40,40,40)$, respectively, except for the original configuration of hidden units. All other setups are identical to the aforementioned experiments, except for the step size or learning rate are fixed as 0.001. Based on the results in Table \ref{Table2hiddens}, the performance for our proposed methods will be improved when the width of hidden layers increases from 20 to 40, but it will degrade for 50. Moreover, their performance are unstable for the cases of three hidden layers. This phenonmen reveals that the ability of Baysian neural networks will become degenerated when the network parameters exceed the requirement\cite{john2010elements}.

\vspace{-0.2cm}
\begin{table}[H]
	\centering
	\caption{REL of different depth and width for hidden units when the above methods is used to solve Example \ref{multiscale_elliptic1d:PDE2}.}
	\vspace{-0.2cm}
	\label{Table2hiddens}
	\begin{tabular}{|l|c|c|c|c|c|c|c|}
		\hline
		Network size     &$(20,20)$ &$(30,30)$ &(40, 40)& (50, 50) &$(20,20,20)$&$(30,30,30)$ &$(40,40,40)$   \\ \hline
	    BPINN\_HMC       & ---      &  ---     & ---    &---       &---         &---          & ---  \\
	    FF\_MBPINN\_HMC  & ---      &---       &---     &---       &---         &---          & ---    \\
	    2FF\_MBPINN\_HMC &---       & ---      &---     &---       &---         &---          & ---    \\      \hline
	    BPINN\_SGD       &0.3983    &0.3964    &0.3977  &0.3962    &0.3959      &0.3968       &0.3946 \\
	    FF\_MBPINN\_SGD  &0.1332    &0.0323    &0.0268  &0.0964    &0.0232      &0.0463       &0.0213        \\
	    2FF\_MBPINN\_SGD &0.0522    &0.0201    &0.0188  &0.0419    &0.0329      &0.0218       &0.0314        \\
		\hline
	\end{tabular}
\vspace{-0.2cm}
\end{table}

\emph{Influence of the number for noise observations:} We study the influence of noise observations for our proposed methods to solve the above two-scale elliptic problem when the $\varepsilon=0.1$ and the noise level is 0.05. Except for the existed 100 noisy observations for solution and force side, we further randomly sample 73 and 148 inner observations, respectively, and 2 boundary observations for solution, we then disturb them by the mean-zero Gaussion noise. As for the force side, 75 and 150 observations sampled randomly from $\Omega$ are also disturbed with mean-zero Gaussian distributed noise. The other setups are identical to the above experiments. Based on the results in Table \ref{Table:solu2multiscale0p05} and Tables \ref{Table2study_sampling75:solu2multiscale0p05} -- \ref{Table2study_sampling150:solu2multiscale0p05}, we observe that the performance of BPINN\_HMC keeps unchanged with the number of noisy observation increasing when it works successully, but the FF\_MBPINN\_SGD and 2FF\_MBPINN\_SGD  will be improved.

\vspace{-0.2cm}
\begin{table}[H]
	\centering
	\caption{The performance of different methods to solve Example \ref{multiscale_elliptic1d:PDE2} for 75 noise observations.}
	\vspace{-0.2cm}
	\begin{tabular}{|l|ccccccc|}
		\hline
		Initial step size(lr)& 0.01  & 0.005 & 0.001 & 0.0005  & 0.0001 & 0.00005 & 0.00001\\ \hline
		BPINN\_HMC           & ---   & ---   & ---   & ---     &---     &0.4011   &0.4054\\
		FF\_MBPINN\_HMC      & ---   & ---   & ---   & ---     &---     &---      &--- \\
		2FF\_MBPINN\_HMC     & ---   & ---   & ---   & ---     &---     &---      &--- \\ \hline
		BPINN\_SGD           &        &      &       &         &        &        &   \\ 
		FF\_MBPINN\_SGD      &0.1789 &0.1766 &0.0554 &0.0628   &0.0673  &0.0698   &0.1145 \\
		2FF\_MBPINN\_SGD     &0.2165 &0.0773 &0.0537 &0.0457   &0.0535  &0.0536   &0.1055\\ \hline
	\end{tabular}
	\label{Table2study_sampling75:solu2multiscale0p05}
\end{table}

\vspace{-0.5cm}
\begin{table}[H]
	\centering
	\caption{The performance of different methods to solve Example \ref{multiscale_elliptic1d:PDE2} for 150 noise observations.}
	\vspace{-0.2cm}
	\begin{tabular}{|l|ccccccc|}
		\hline
		Initial step size(lr)& 0.01  & 0.005 & 0.001 & 0.0005  & 0.0001 & 0.00005 & 0.00001\\ \hline
		BPINN\_HMC           & ---   & ---   & ---   & ---     &---     &0.4013   &0.4023\\
		FF\_MBPINN\_HMC      & ---   & ---   & ---   & ---     &---     &---      &--- \\
		2FF\_MBPINN\_HMC     & ---   & ---   & ---   & ---     &---     &---      &--- \\ \hline
		BPINN\_SGD           &        &      &       &         &        &        &   \\ 
		FF\_MBPINN\_SGD      &0.0825 &0.0610 &0.0499 &0.0398   &0.0494  &0.0592   &0.9216 \\
		2FF\_MBPINN\_SGD     &0.1801 &0.1205 &0.0141 &0.0194   &0.0203  &0.0208   &0.1015   \\ \hline
	\end{tabular}
	\label{Table2study_sampling150:solu2multiscale0p05}
	\vspace{-0.2cm}
\end{table}

\begin{example}\label{NonlinearPoisson:PDE1}
Now, let us study the prtformance for our proposed MBPINN method to the general peoblems. We consider the following non-linear Poisson problem with two frequency components in $\Omega=[0,1]$: 
\vspace{-0.1cm}
\begin{equation}\label{NonLinearPoisson1D}
    \begin{cases}
        0.01\Delta u(x) + k(x) u(x) = f(x)\\
        u(0) = u(1) = 0.1.
    \end{cases}
\end{equation}
with a coefficient term $k(x)$ as follows 
\vspace{-0.1cm}
\begin{equation}
k(x) = 0.1 + \exp\left(-0.5 \frac{(x - 0.5)^2}{0.15^2}\right)
\end{equation}
An exact solution is given by 
\vspace{-0.1cm}
\begin{equation}
u_{\text{true}}(x) = \sin(2\pi x) + 0.1\cos(10\pi x)
\end{equation}
it naturally induces the force term 
\vspace{-0.1cm}
\begin{equation}
    f(x) =  0.01 \left(-4\pi^2\sin(2\pi x) - 10\pi^2\cos(10\pi x)\right) + k(x) \cdot u(x).
\end{equation}
\end{example}

\begin{figure} [H]
	\centering
	\vspace{-0.1cm}   
	\subfigure[BPINN\_HMC] {
		\label{BPINN_Solu2NonLinPoi0p05}     
		\includegraphics[scale=0.275]{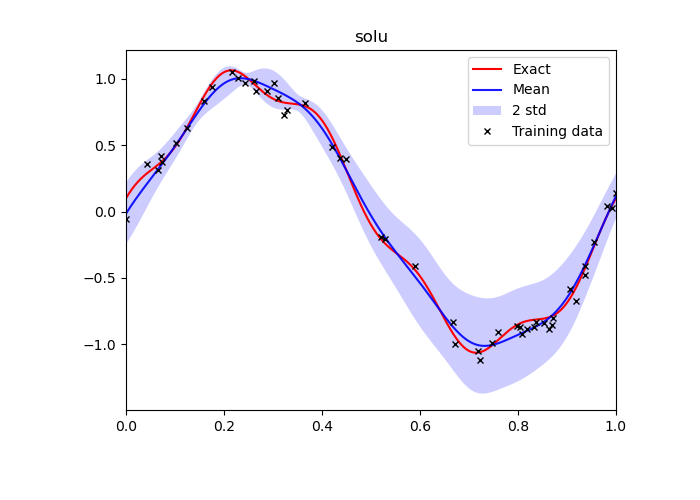}  
	}
\vspace{-0.1cm}
	\subfigure[FF\_MBPINN\_HMC] {
		\label{FF_MBPINN_Solu2NonLinPoi0p05}     
		\includegraphics[scale=0.275]{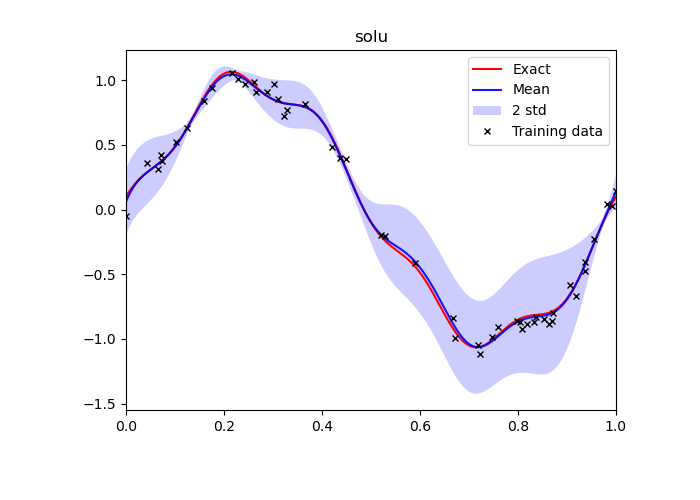}  
	}  
\vspace{-0.1cm}       
	\subfigure[2FF\_MBPINN\_HMC] {
		\label{2FF_MBPINN_Solu2NonLinPoi0p05}     
		\includegraphics[scale=0.275]{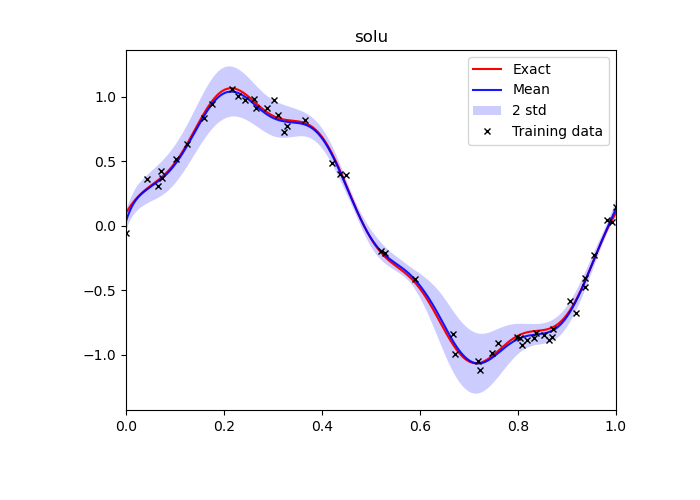}  
	} 
\vspace{-0.1cm} 
	\subfigure[BPINN\_SGD] {
		\label{BPINNSGD_Solu2NonLinPoi0p05}     
		\includegraphics[scale=0.275]{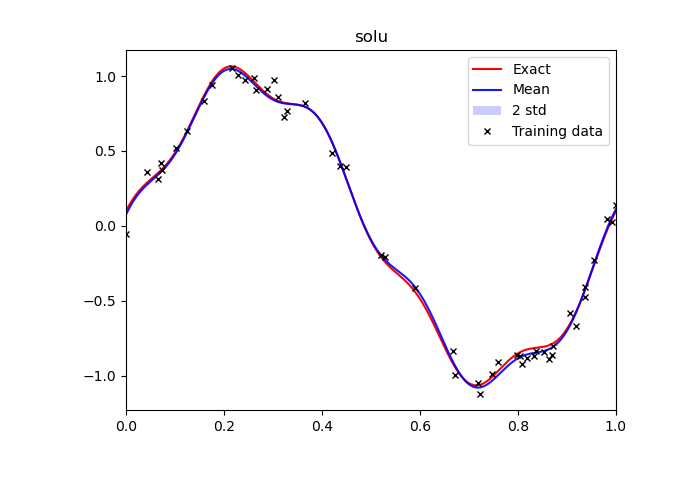}  
	} 
\vspace{-0.1cm} 
	\subfigure[FF\_MBPINN\_SGD] {
		\label{SGD_FF_MBPINN_Solu2NonLinPoi0p05}     
		\includegraphics[scale=0.275]{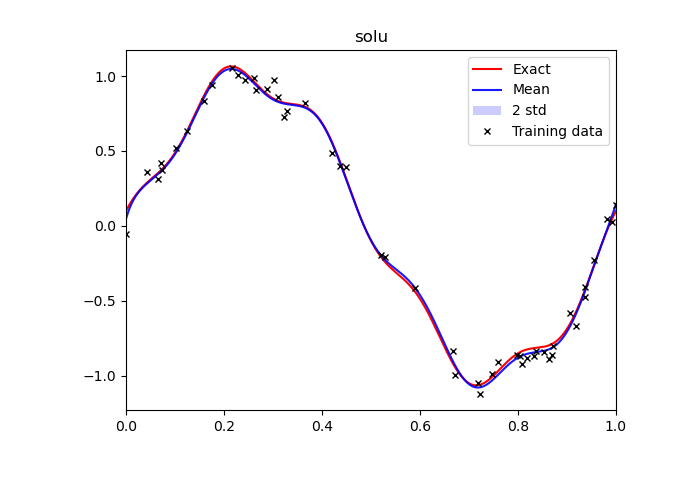}  
	} 
\vspace{-0.1cm}        
	\subfigure[2FF\_MBPINN\_SGD] {
		\label{SGD_2FF_MBPINN_Solu2NonLinPoi0p05}     
		\includegraphics[scale=0.275]{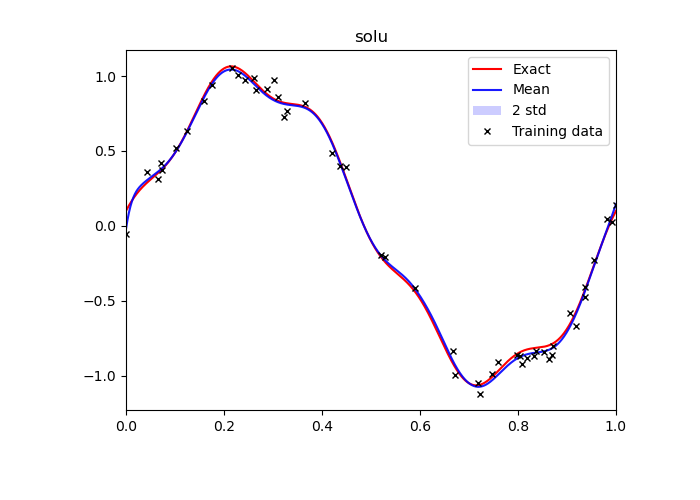}  
	}
	\caption{Visualization of the solution for different methods to solve Example \ref{NonlinearPoisson:PDE1} when the noise level is 0.05.} 
	\label{Solu2NonlinPoisson_0p5}   
	\vspace{-0.5cm}      
\end{figure}

\begin{figure} [H]
	\centering   
	\subfigure[BPINN\_HMC] {
		\label{BPINN_Para2NonLinPoi0p05}     
		\includegraphics[scale=0.275]{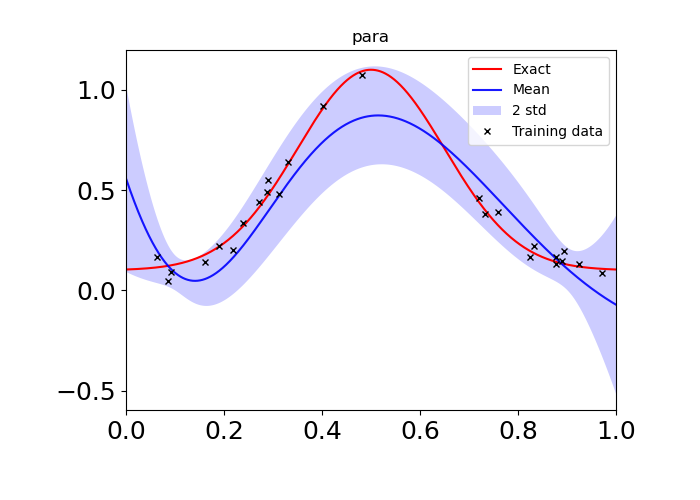}  
	} 
	\subfigure[FF\_MBPINN\_HMC] {
		\label{FF_MBPINN_Para2NonLinPoi0p05}     
		\includegraphics[scale=0.275]{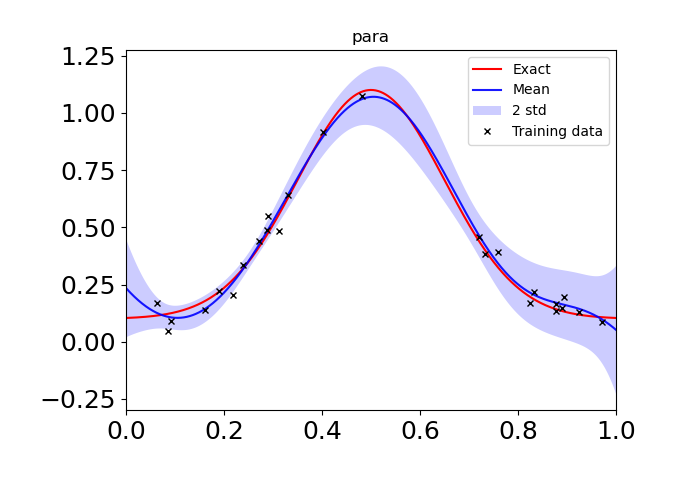}  
	}          
	\subfigure[2FF\_MBPINN\_HMC] {
		\label{2FF_MBPINN_Para2NonLinPoi0p05}     
		\includegraphics[scale=0.275]{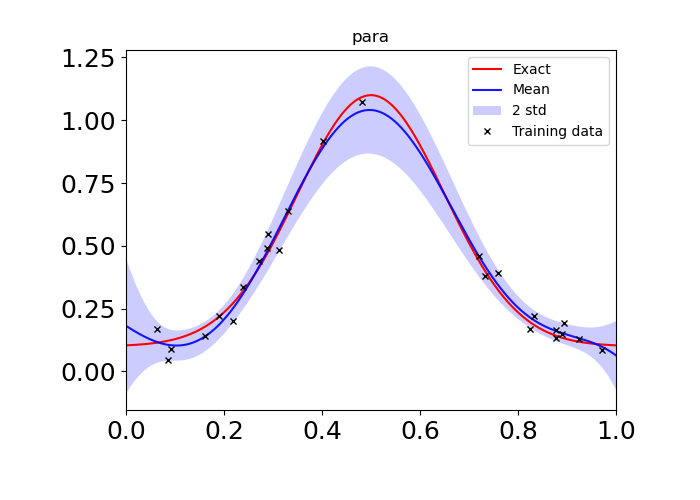}  
	} 
	\subfigure[BPINN\_SGD] {
		\label{BPINNSGD_Para2NonLinPoi0p05}     
		\includegraphics[scale=0.275]{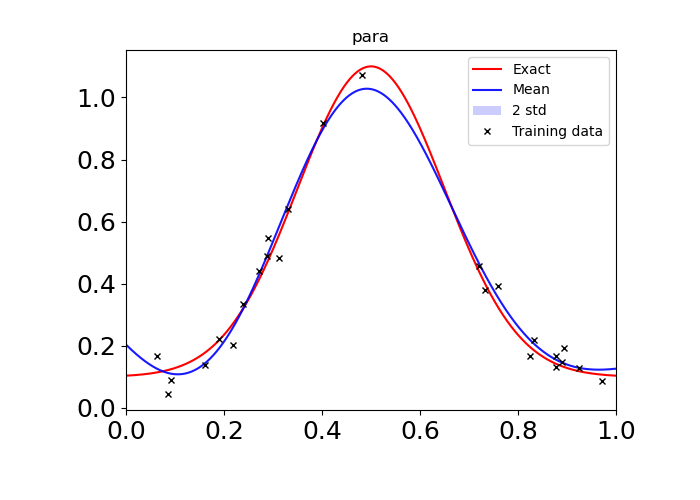}  
	} 
	\subfigure[FF\_MBPINN\_SGD] {
		\label{SGD_FF_MBPINN_Para2NonLinPoi0p05}     
		\includegraphics[scale=0.275]{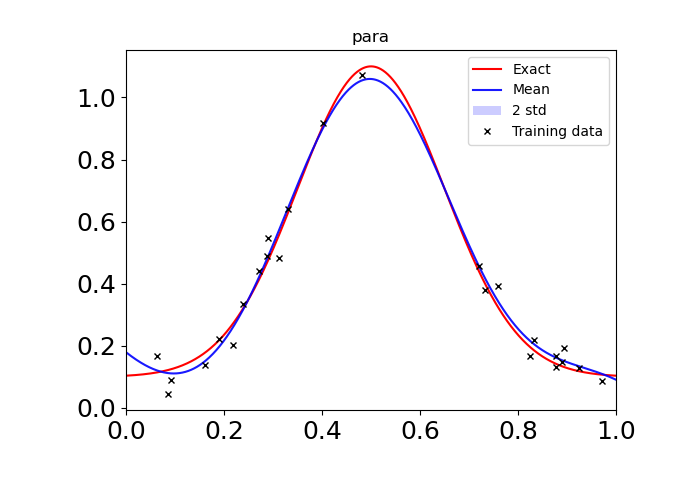}  
	}          
	\subfigure[2FF\_MBPINN\_SGD] {
		\label{SGD_2FF_MBPINN_Para2NonLinPoi0p05}     
		\includegraphics[scale=0.275]{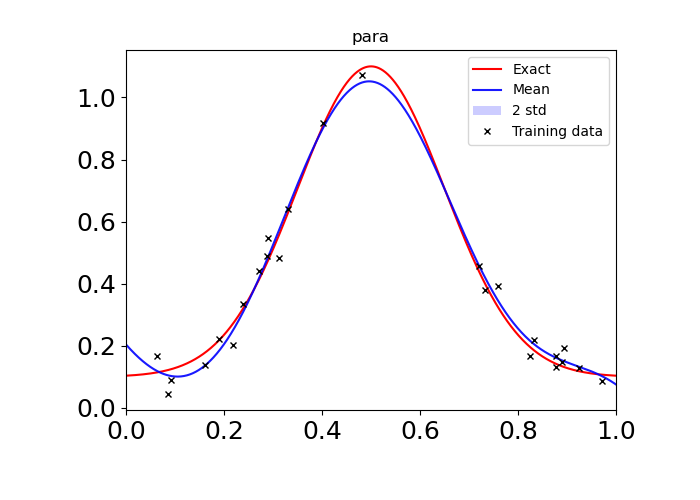}  
	} 
\vspace{-0.25cm}  
	\caption{Visualization of the parameter for different methods to solve Example \ref{NonlinearPoisson:PDE1} when the noise level is 0.05.} 
	\label{Para2NonlinPoisson_0p5}    
	\vspace{-0.15cm}       
\end{figure}

We solve the above problem by employing the aforementioned six methods that their solvers have two hidden layers and each layer has 30 units. In this example, 50 observations sampled randomly from $\Omega$ for the solution and force side, respectively, are disturbed with mean-zero Gaussian distributed noise. In addition, the 25 observations sampled randomly from $\Omega$ for the coefficient term, $k(x)$, are disturbed as well. To recover the solution and coefficient term simultaneously, two ansatzes expressed by DNN are used in  BPINN method, and two ansatzes expressed by MscaleDNN with FFM pipelines are used in other methods. The standard deviation for $\Lambda$ of FF\_MBPINN is set as 5 and 0.5 for solution and coefficient, respectively.  The standard deviations for $\Lambda_1$ and $\Lambda_2$ of 2FF\_MBPINN used to expressed solution are set as 1 and 5, but for coefficient are set as 0.1 and 0.5, respectively. Additionally, the SNRs for solution, coefficient term and force side are 22.3147, 20.2577 and 23.2779, respectively, when the noise level is 0.05. Finally, we list and depict the related experiment results in Tables \ref{solu2NonLinPoisson0p05} -- \ref{para2NonLinPoisson0p05} and Figures \ref{Solu2NonlinPoisson_0p5} -- \ref{Para2NonlinPoisson_0p5}, respectively. 


\vspace{-0.2cm}
\begin{table}[H]
    \centering
    \caption{REL of solution for different methods to solve Example \ref{NonlinearPoisson:PDE1} when the noise level is 0.05.}
    \vspace{-0.2cm}
    \begin{tabular}{|l|ccccccc|}
    \hline 
Initial step size(lr)& 0.01   & 0.005  & 0.001  & 0.0005  & 0.0001 & 0.00005 & 0.00001\\ \hline
    BPINN\_HMC       & ---    & ---    & ---    & ---     &0.0715  & 0.1176 & 0.6259 \\
    FF\_MBPINN\_HMC  & ---    & ---    & ---    & ---     &0.0253  &0.0228  &0.0559\\
    2FF\_MBPINN\_HMC & ---    & ---    & ---    & ---     &0.0261  &0.0305  &0.0626  \\ \hline
    BPINN\_SGD       &0.0279  &0.0266  &0.0248  &0.0234   &0.2940  &0.1133  &0.6066  \\
    FF\_MBPINN\_SGD  &0.0251  &0.0247  &0.0238  & 0.0249  &0.0248  &0.0250  &0.0285        \\
    2FF\_MBPINN\_SGD &0.0257  &0.0276  &0.0261  &0.0261   &0.0260  &0.0252  &0.0292     \\ \hline
    \end{tabular}
\label{solu2NonLinPoisson0p05}
\vspace{-0.4cm}
\end{table}

\begin{table}[H]
	\centering
	\caption{REL of parameter for different methods to solve Example \ref{NonlinearPoisson:PDE1} when the noise level is 0.05.}
	\vspace{-0.2cm}
	\begin{tabular}{|l|ccccccc|}
		\hline 
	Initial step size(lr)& 0.01   & 0.005  & 0.001  & 0.0005  & 0.0001 & 0.00005& 0.00001\\ \hline
		BPINN\_HMC       & ---    & ---    & ---    & ---     &0.2361  &0.3379  & 0.8291 \\
		FF\_MBPINN\_HMC  & ---    & ---    & ---    & ---     &0.0482  &0.0679  &0.2688 \\
		2FF\_MBPINN\_HMC & ---    & ---    & ---    & ---     &0.0491  &0.0985  &0.3334 \\ \hline
		BPINN\_SGD       &0.0668  & 0.0410 &0.0605  &0.0487   &1.6010  &0.4244  &0.6420\\
		FF\_MBPINN\_SGD  &0.0477  &0.0394  &0.0380  &0.0379   &0.0380  &0.0375  &0.0942      \\
		2FF\_MBPINN\_SGD &0.0422  &0.0510  &0.0476  &0.0478   &0.0457  &0.0466  &0.1013      \\ \hline
	\end{tabular}
	\label{para2NonLinPoisson0p05}
\end{table}

From the results in Tables~ \ref{solu2NonLinPoisson0p05} -- \ref{para2NonLinPoisson0p05} and Figs. \ref{Solu2NonlinPoisson_0p5} -- \ref{Para2NonlinPoisson_0p5}, the FF\_MBPINN and 2FF\_MBPINN method are superior to the classical BPINN in terms of the same step szie or learning rate.  In which, BPINN\_HMC method fails to work when the step size is relatively large and its performance will be deteriorated with the noise level increasing when it is successful to work. Although the MBPINN\_HMC and 2FF\_MBPINN\_HMC methods also face the issues with failure, they perform better when they succeeds in working. It is important to note that a plus of the MBPINN\_HMC method is its ability to provide interval estimates for parameters, whereas MBPINN\_SGD only offers point estimates. Moreover, the SGD driven BPINN, FF\_MBPINN and 2FF\_MBPINN outperform the HMC driven ones, and they can provide the satifactory point estimation for the interval parameters. Additionally, the performance of FF\_MBPINN and 2ff\_MBPINN mthod are completed. In conclusion, the MBPINN\_SGD method consistently provides the best point estimates for parameters and reliably produces results. 

\subsection{Performance of MBPINN for Solving two-dimensional PDEs}

\begin{example}\label{Multiscale2D_Epsilon}
	Now, our goal is to deal with the following multiscale elliptic equation in a regular rectangle domain $\Omega=[0, 1]\times[0, 1]$. 
	\begin{equation}\label{eq:Multiscale2D_Epsilon}
	\begin{cases}
	\displaystyle -\textup{div}\left(A^{\varepsilon}(x_1,x_2)\nabla u^{\varepsilon}(x_1, x_2)\right) = f(x_1,x_2)\\
	\displaystyle  u^{\varepsilon}(x_1, 0) = \frac{1}{4}x_1^4+\frac{\varepsilon}{16}x_1^2\sin\left(\frac{2\pi x_1^2}{\varepsilon}\right)+\frac{\varepsilon^2}{32\pi^2}\cos\left(\frac{2\pi x_1^2}{\varepsilon}\right)\\
	\displaystyle  u^{\varepsilon}(x_1, 1) = \frac{1}{4}(x_1^2+1)^2+\frac{\varepsilon}{16}(x_1^2+1)\sin\left(\frac{2\pi(1+ x_1^2)}{\varepsilon}\right)+\frac{\varepsilon^2}{32\pi^2}\cos\left(\frac{2\pi (x_1^2+1)}{\varepsilon}\right)\\
	\displaystyle  u^{\varepsilon}(0, x_2) = \frac{1}{4}x_2^4+\frac{\varepsilon}{16}x_2^2\sin\left(\frac{2\pi x_2^2}{\varepsilon}\right)+\frac{\varepsilon^2}{32\pi^2}\cos\left(\frac{2\pi x_2^2}{\varepsilon}\right)\\
	\displaystyle  u^{\varepsilon}(1, x_2) = \frac{1}{4}(1+x_2^2)^2+\frac{\varepsilon}{16}(1+x_2^2)\sin\left(\frac{2\pi(1+ x_2^2)}{\varepsilon}\right)+\frac{\varepsilon^2}{32\pi^2}\cos\left(\frac{2\pi (1+x_2^2)}{\varepsilon}\right)
	\end{cases}
	\end{equation}
	in which 
	\begin{equation}
	\displaystyle A^\varepsilon(x_1, x_2) = \frac{1}{4 + \cos\left(\frac{2\pi (x_1^2+x_2^2)}{\varepsilon}\right)},
	\end{equation}
	with $\varepsilon > 0$ being a small constant and $f(x_1,x_2)=-(x_1^2+x_2^2)$. Under these conditions, an exact solution is given by
	\begin{equation}
	u^{\varepsilon}(x_1,x_2) = \frac{1}{4}(x_1^2+x_2^2)^2+\frac{\varepsilon}{16\pi}(x_1^2+x_2^2)\sin\left(\frac{2\pi(x_1^2+ x_2^2)}{\varepsilon}\right)+\frac{\varepsilon^2}{32\pi^2}\cos\left(\frac{2\pi (x_1^2+x_2^2)}{\varepsilon}\right).
	\end{equation}
\end{example}

We utilize the aforementioned six methods to solve the above two-dimensional multiscale elliptic problem with $\varepsilon=0.5$ when the niose level is 0.01 and 0.02, respectively. Their solvers all have two hidden layers and each layer has 40 neuron units. In this example,  we randomly sample 500 observations for force side in $\Omega$ and disturb them by Gaussian distribution noise with zero mean error. In addition, 500 random observations composed of 400 interior points and 100 boundary points for solution in $\Omega$ are disturbed as well. To recover the solution from these noise observations, an ansatze expressed by DNN is used in BPINN method and an ansatze expressed by MscaleDNN with FFM pipelines is used in other methods. The standard deviation of $\Lambda$ for FF\_MBPINN is set as 5 and the standard deviations of $\Lambda_1$ and $\Lambda_2$ for 2FF\_MBPINN are set as 1 and 5, respectively. We list and plot the related experiment results in Tables \ref{snr2multiscale_elliptic2d} -- \ref{solu2Multiscale2D_Eps0p5_Noise0p02} and Figure \ref{Results2Multiscale2D_Eps0p5}, respectively.

\vspace{-0.2cm}
\begin{table}[!ht]
	\centering
	\caption{SNR of solution and force side under various noise level for  Example \ref{Multiscale2D_Epsilon}.}
	\vspace{-0.2cm}
	\label{snr2multiscale_elliptic2d}
	\begin{tabular}{|l|c|c|}
		\hline
		Noise level      &$0.01$           &$0.02$          \\  \hline
		solution	     &28.117           &22.763       \\ \hline
		force side       &37.963           &31.653       \\  \hline
	\end{tabular}
\vspace{-0.5cm}
\end{table}

\begin{table}[H]
	\centering
	\caption{REL of solution for different methods to solve Example \ref{Multiscale2D_Epsilon} when the noise level is 0.01.}
	\vspace{-0.15cm}
	\begin{tabular}{|l|ccccccc|}
		\hline 
		Initial step size(lr)& 0.01   & 0.005  & 0.001  & 0.0005  & 0.0001 & 0.00005 & 0.00001\\ \hline
		BPINN\_HMC           & ---    & ---    & ---    & ---     & 0.7008 & 0.9686  & 1.0734 \\
		FF\_MBPINN\_HMC      & ---    & ---    & ---    & ---     & ---    & ---     & --- \\
		2FF\_MBPINN\_HMC     & ---    & ---    & ---    & ---     & ---    & ---     & --- \\ \hline
		  BPINN\_SGD         &0.0530  &0.0733   &0.8672 &0.9507   &0.9662  &0.9673   &1.0768\\
		FF\_MBPINN\_SGD      &0.1329  &0.0426  &0.0748  &0.0889   &0.1323  & 0.1537  &0.4266      \\
		2FF\_MBPINN\_SGD     &0.0690  &0.0201  &0.1441  &0.1887   &0.2375  &0.2596   &0.4843       \\ \hline
	\end{tabular}
	\label{solu2Multiscale2D_Eps0p5_Noise0p01}
	\vspace{-0.4cm}
\end{table}

\begin{table}[H]
	\centering
	\caption{REL of solution for different methods to solve Example \ref{Multiscale2D_Epsilon} when the noise level is 0.02.}
	\vspace{-0.15cm}
	\begin{tabular}{|l|ccccccc|}
		\hline 
		Initial step size(lr)& 0.01   & 0.005  & 0.001  & 0.0005  & 0.0001 & 0.00005 & 0.00001\\ \hline
		BPINN\_HMC           & ---    & ---    & ---    & ---     &0.9136  &0.9278   &0.8566  \\
		FF\_MBPINN\_HMC      & ---    & ---    & ---    & ---     & ---    & ---     & --- \\
		2FF\_MBPINN\_HMC     & ---    & ---    & ---    & ---     & ---    & ---     & --- \\ \hline
			BPINN\_SGD       &0.0397  &0.0847   &0.6415 &0.8368   &0.9082  &0.9089   &0.9755\\
		FF\_MBPINN\_SGD      &0.0481  &0.0351  & 0.0863 &0.0941   &0.1250  &0.1470   &0.4315  \\
		2FF\_MBPINN\_SGD     &0.1189  &0.0299  &0.1391  &0.1743   &0.2075  &0.2233   &0.4471            \\ \hline
	\end{tabular}
	\label{solu2Multiscale2D_Eps0p5_Noise0p02}
	\vspace{-0.2cm}
\end{table}

\vspace{-0.4cm}
\begin{figure}[H]
	\centering  
	\subfigure[$A^{\varepsilon}(x_1,x_2)$]{
		\label{Multiscale2D:coef}     
		\includegraphics[scale=0.425]{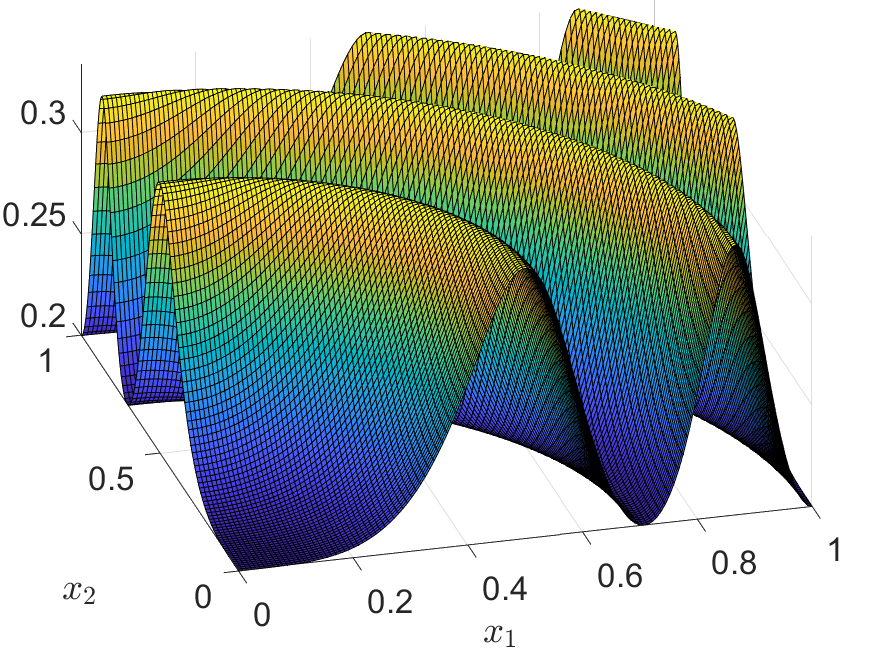}  
	}  
	\subfigure[Exact solution and noisy observations]{
		\label{Multiscale2D:solu}     
		\includegraphics[scale=0.425]{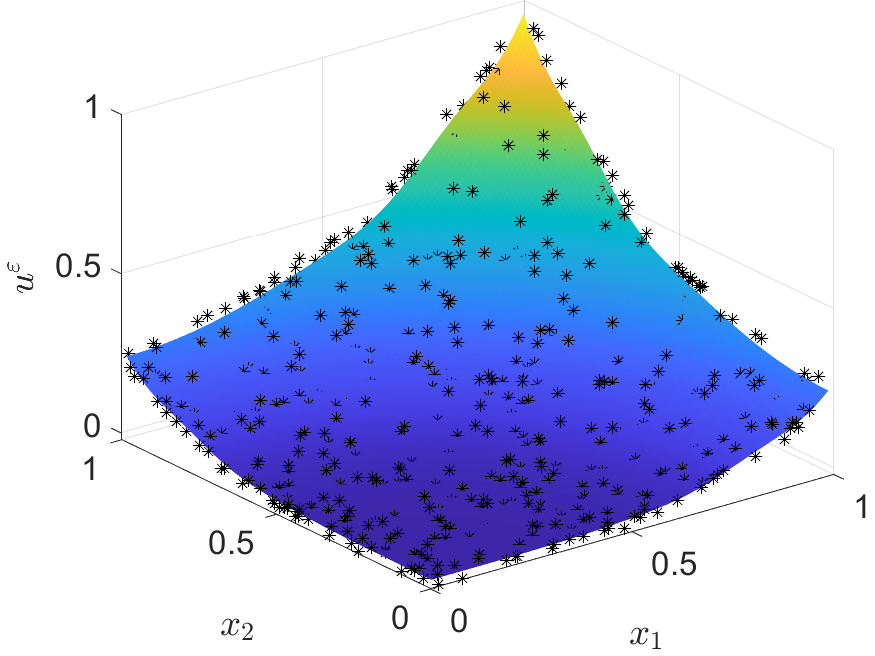}  
	} 
\vspace{-0.3cm}
\caption{Coefficient and exact solution with noisy observations for Example \ref{Multiscale2D_Epsilon}.} 
\vspace{-0.4cm}
\end{figure}

\begin{figure}[H]
	\centering  
	\subfigure[BPINN\_HMC for step size=0.0005]{
		\label{BPINN_HMC2Multiscale2D:PointERR}     
		\includegraphics[scale=0.425]{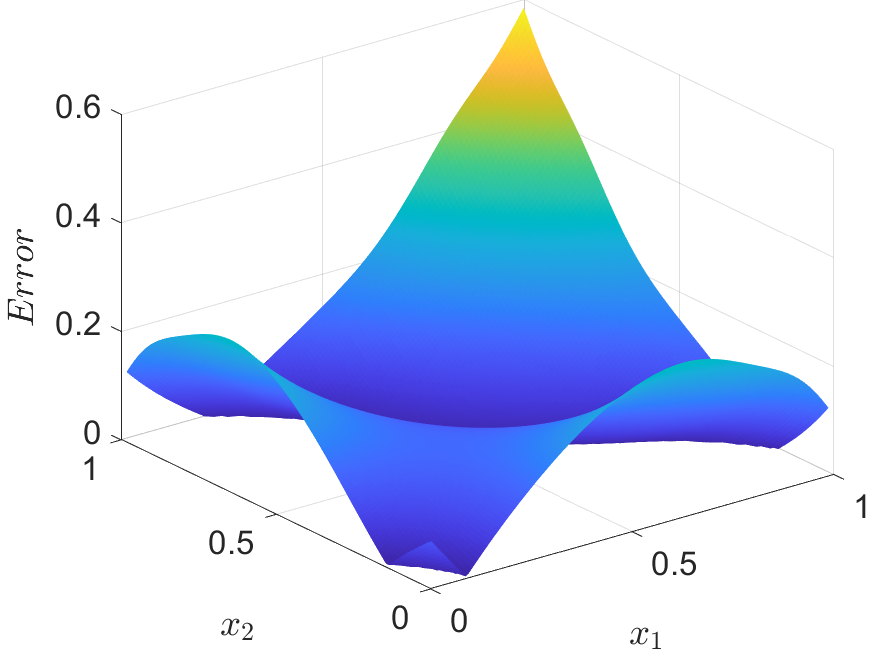}    
	}
	\subfigure[MBPINN\_SGD for lr=0.005]{
		\label{BPINN_SGD2Multiscale2D:PointERR}     
		\includegraphics[scale=0.425]{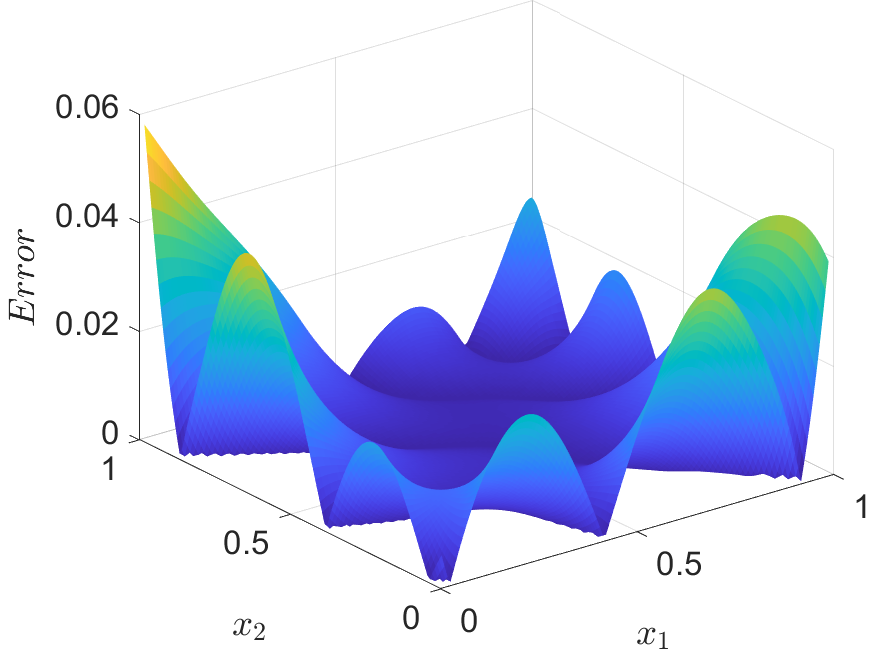}     
	} 
	\subfigure[FF\_MBPINN\_SGD for lr=0.005]{
		\label{FF_BPINN_SGD2Multiscale2D:PointERR}     
		\includegraphics[scale=0.425]{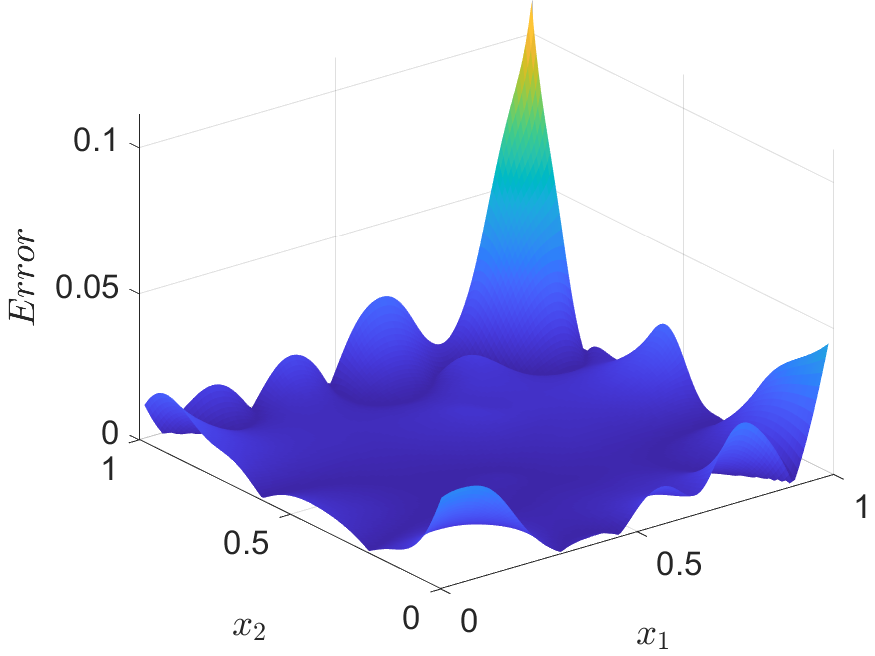}     
	} 
	\subfigure[2FF\_MBPINN\_SGD for lr=0.005]{
		\label{2FF_BPINN_SGD2Multiscale2D:PointERR}     
		\includegraphics[scale=0.425]{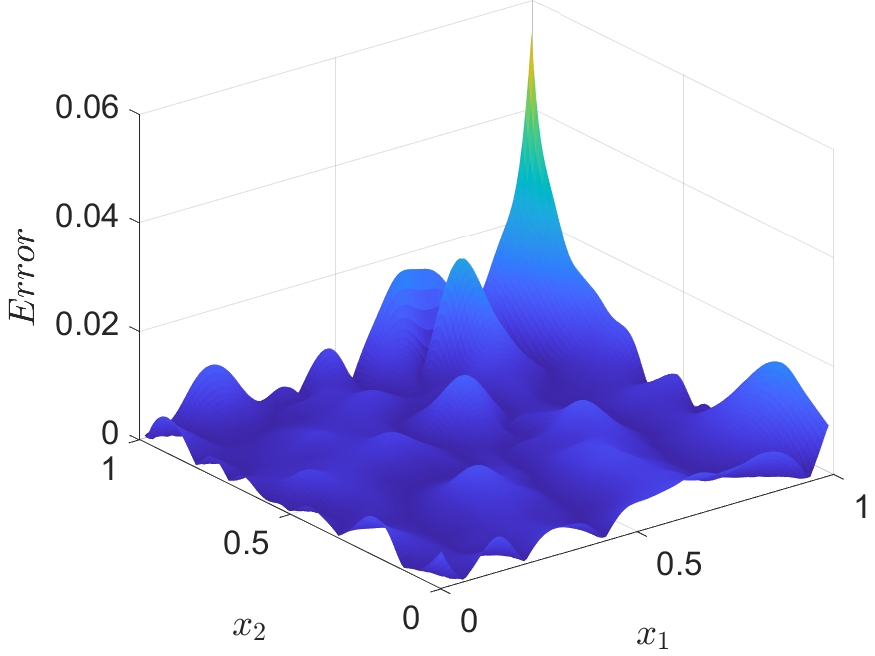}     
	}  
\vspace{-0.3cm}
	\caption{The point-wise error of numerical methods for Example \ref{Multiscale2D_Epsilon} when the noise level is 0.01} 
	\label{Results2Multiscale2D_Eps0p5}    
	\vspace{-0.3cm}     
\end{figure}

Based the point-wise errors in Figure \ref{Results2Multiscale2D_Eps0p5} and the relative error in Tables \ref{snr2multiscale_elliptic2d} -- \ref{solu2Multiscale2D_Eps0p5_Noise0p02}, we can conclude that the both FF\_MBPINN\_SGD and 2FF\_MBPINN\_SGD outperform the other methods, especially at smaller step sizes, when these methods are successful to work. It is important to note that the HMC driven MBPINN methods are not able to provide the solution for our problem. The BPINN\_HMC method  works and provides a result under the small step size, but its performance is terrible. This indicates that the result was entirely dependent on the initialization and their performance is  affected by the step size used to update the interval parameters of solver, in particularly for multi-scale PDEs. This highlights a critical limitation of this method for the utilization to the multi-scale scenario and indicates the superiority of SGD technique. 
%

\begin{example}\label{LinPoisson2D}
Now, our goal is to deal with the following general Poisson equation in a regular rectangle domain $\Omega=[-1, 1]\times[-1, 1]$. 
\begin{equation}\label{eq:LinearPoisson3}
    \begin{cases}
        \Delta u(x_1, x_2) = f(x_1,x_2)\\
        u(-1, y) = u(1, y) = e^{\sin(\pi x_2)}\\
        u(x, -1) = u(x, 1) = e^{\sin(\pi x_1)}.
    \end{cases}
\end{equation}
An exact solution is given by
\begin{equation}
u(x_1,x_2) = e^{\sin(\pi x_1)}e^{\sin(\pi x_2)}
\end{equation}
such that 
\begin{equation*}
    f(x_1,x_2)=\pi^2e^{\sin(\pi x_1)}e^{\sin(\pi x_2)}[\cos^2(\pi x_1)-\sin(\pi x_1)+\cos^2(\pi x_2)-\sin(\pi x_2)]
\end{equation*}
\end{example}

We solve the above smooth Poisson problem by employing the aforementioned six methods when the noise level is 0.2. In this example, we randomly sample 800 random observations for force side and solution in $\Omega$, respectively.  We then disturb these observations by Gaussian distributed noise with zero mean error. In addition, the standard deviation of $\Lambda$ for FF\_MBPINN is set as 1.0 and the standard deviations of $\Lambda_1$ and $\Lambda_2$ for 2FF\_MBPINN are set as 0.2 and 1.0, respectively.  The all setups are same as the above Example \ref{Multiscale2D_Epsilon}. In our test, the SNRs for solution and force side are 20.538 and 43.587, respectively. We list and depict the related experiment results in Table \ref{Table:Linear_Poisson2D_N0.05} and Figure \ref{Results2Poisson2D}, respectively.
\vspace{-0.2cm} 
\begin{table}[H]
    \centering
    \caption{REL of different method with two hidden layers to solve Example \ref{LinPoisson2D} when the noise is 0.2.}
     \vspace{-0.2cm} 
    \begin{tabular}{|l|ccccccc|}
    \hline
Initial step Size(lr)&0.01  & 0.005 & 0.001 & 0.0005  & 0.0001 & 0.00005 & 0.00001\\ \hline
    BPINN\_HMC       & ---  & ---   & ---   & ---     & ---    &0.0107   &0.0343 \\
    FF\_MBPINN\_HMC  & ---  & ---   & ---   & ---     & ---    &---      &0.0294  \\
    2FF\_MBPINN\_HMC & ---  & ---   & ---   & ---     & ---    &---      &0.0381  \\ \hline
    BPINN\_SGD       &0.0130&0.0127 &0.0203 &0.0186   &0.0272  &0.0497   &0.5835  \\
     FF\_MBPINN\_SGD &0.0102&0.0110 &0.0097 &0.0095   &0.0094  &0.0099   &0.0647  \\
    2FF\_MBPINN\_SGD &0.0186&0.0121 &0.0105 &0.0106   &0.0116  &0.0131   &0.0613  \\
    \hline
    \end{tabular}
\label{Table:Linear_Poisson2D_N0.05}
\end{table}

 \vspace{-0.25cm} 
\begin{figure}[H]
	\centering   
	\label{2dPoisson:solu}     
	\includegraphics[scale=0.5]{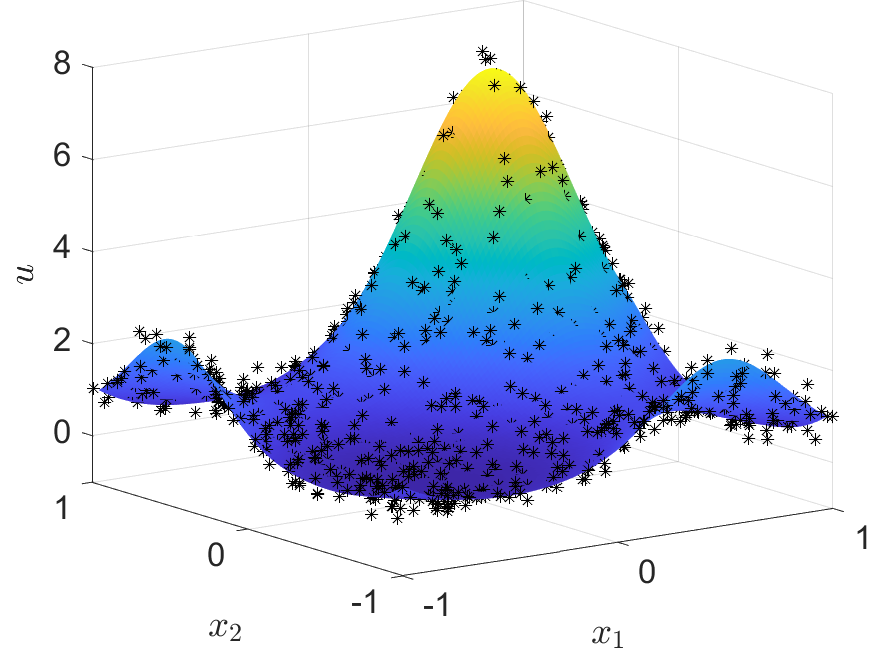}  
	\caption{Exact solution and noisy observations for Example \ref{LinPoisson2D}.}
\end{figure}

\begin{figure}[H]
	\centering   
	\subfigure[BPINN\_HMC for step size=0.00005]{
		\label{BPINN2Poisson2D:PointERR}     
		\includegraphics[scale=0.425]{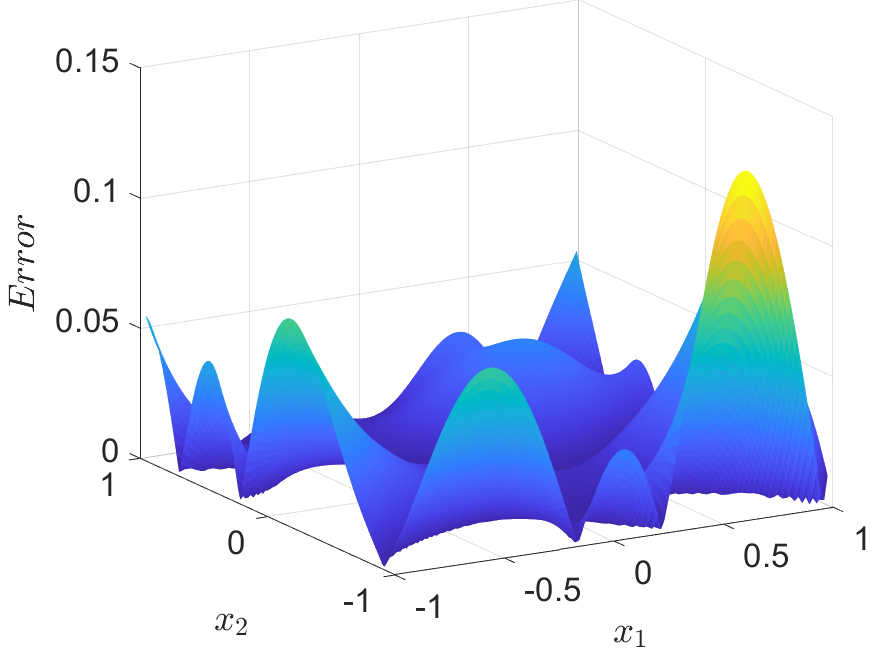}
	}
	\subfigure[FF\_MBPINN\_HMC for step size=0.00001]{
		\label{2FFMBPINNHMC2Poisson2D:PointERR}     
		\includegraphics[scale=0.425]{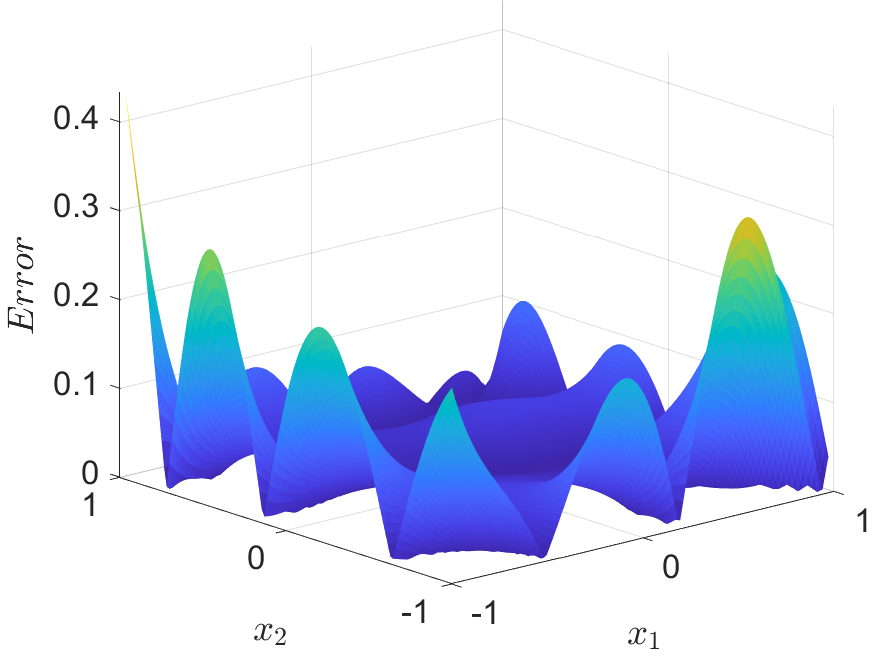}     
	} 
	\subfigure[2FF\_MBPINN\_HMC for step size=0.00001]{
		\label{2FFMBPINN2Poisson2D:PointERR}     
		\includegraphics[scale=0.425]{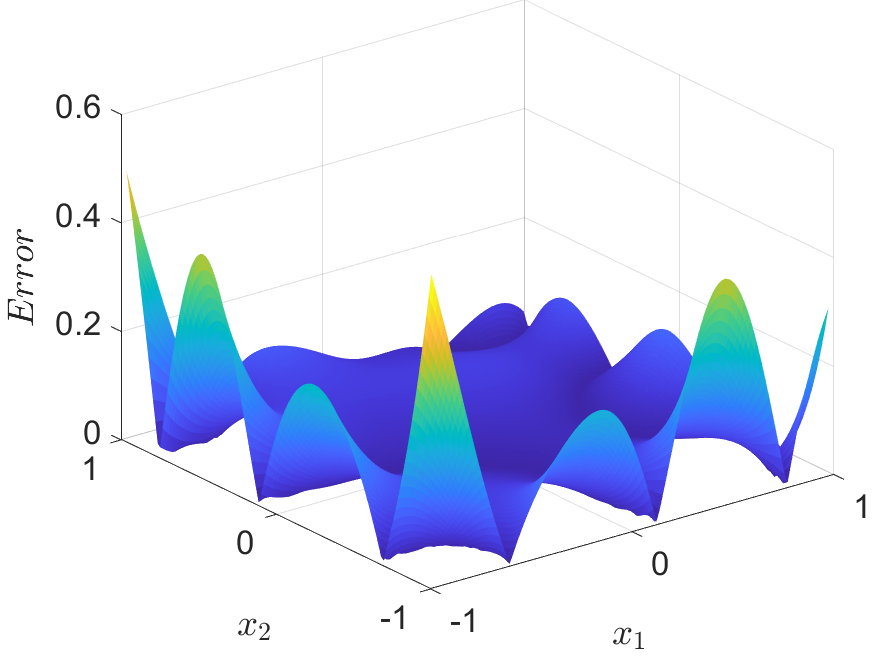}     
	} 
	\subfigure[BPINN\_SGD for lr =0.005]{
		\label{FFMBPINNHMC2Poisson2D:PointERR}     
		\includegraphics[scale=0.425]{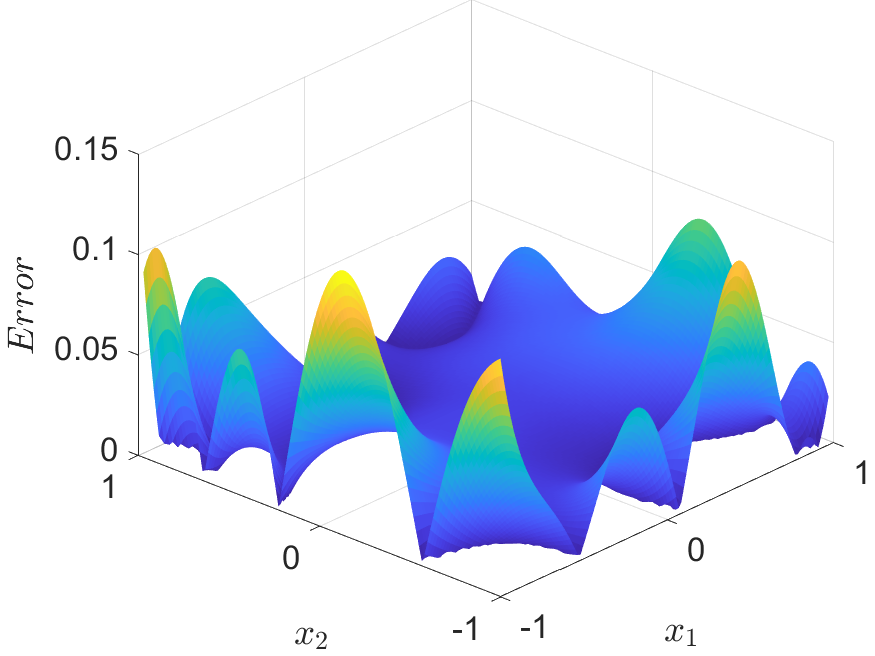}
	} 
	\subfigure[FF\_MBPINN\_SGD for lr =0.0005]{
		\label{FFMBPINNSGD2Poisson2D:PointERR}     
		\includegraphics[scale=0.425]{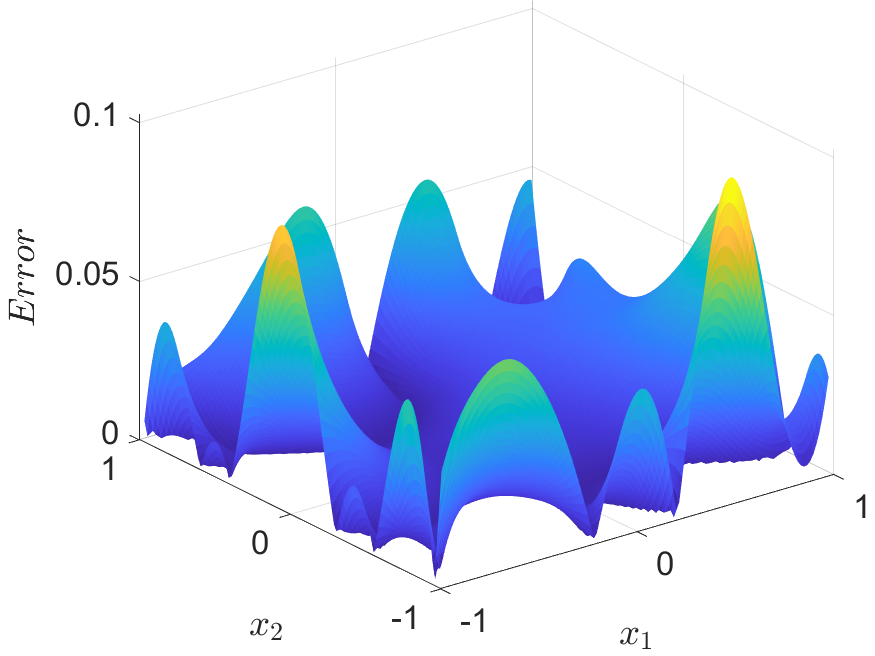}     
	} 
	\subfigure[2FF\_MBPINN\_SGD for lr =0.0005]{
		\label{2FFMBPINNSGD2Poisson2D:PointERR}     
		\includegraphics[scale=0.425]{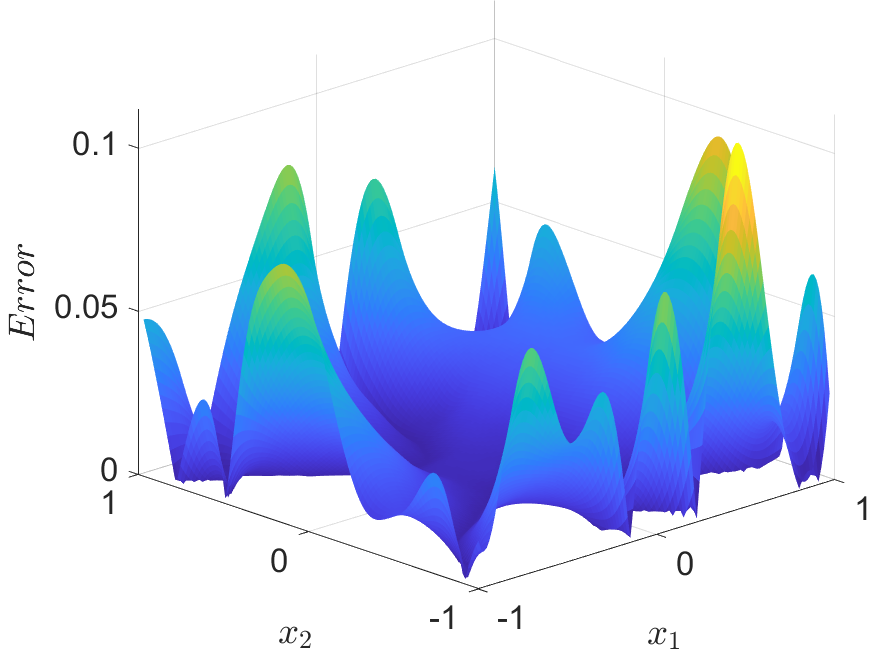}     
	} 
	\caption{The point-wise error of numerical methods for Example 3 when noise level is 0.05} 
	\label{Results2Poisson2D}         
\end{figure}

The results in Table \ref{Table:Linear_Poisson2D_N0.05} and the point wise absolute error in Figure \ref{Results2Poisson2D} indicate that the SGD driven BPINN and MBPINN methods are generally superior to the HMC-based ones when they are successful to solve this problem. Furthermore, the use of SGD proves to be more effective in reducing errors, especially at moderate step sizes, demonstrating that even in two-dimensional settings, SGD-based methods remain effective and robust. It is important to note that SGD-based methods have significantly lower computational costs compared to HMC-based methods, a difference that becomes even more pronounced in two-dimensional problems.
In addition, the failure of convergence for HMC estimation method with large step zise is still occurred when classical BPINN and our proposed method are used to resolve this general Poisson problem in two dimensional space, in particular for our proposed methods. Under this case, our proposed methods still outperform the BPINN method.

\section{Conclusion}\label{sec:conclusion}
The classical BPINN method faces significant challenges in practical applications, particularly due to the convergence issues of the HMC sampling method when solving general elliptic PDEs, which are exacerbated as the complexity of the PDEs increases. Additionally, the basic setup of BPINN is somewhat simplistic and performs poorly when the complexity of PDEs increases and the step size used to update the solver's interval parameters is large. To address these issues simultaneously, we propose an enhanced BPINN by incorporating a FFM induced multi-scale DNN into this architecture, which has been tested and compared against BPINN method on general Poisson across one and two-dimensional spaces with noisy data. Moreover, we reframe the estimation approach for the interval parameters of this proposed novel method in terms of the failure of the HMC caused by the step size and the complexity of PDE problems, especially for multi-scale elliptic PDEs. Our results demonstrate that the SGD driven MBPINN (1) is more robust than HMC, (2) incurs lower computational costs, and (3) offers greater flexibility in handling complex problems. However, when MBPINN is combined with HMC, the convergence issues of HMC worsen due to the inherent complexity of the neural network solver and the problem to be solved, a challenge that warrants further investigation in the future.

\section*{Credit authorship contribution Statement}
Yilong Hou: Methodology, Investigation, Formal analysis, Validation, Writing - Original Draft. Xi'an Li: Conceptualization, Methodology, Investigation, Formal analysis, Validation, Writing - Review \& Editing. Jinran Wu:  Formal analysis, Writing - Review \& Editing.  You-Gan Wang: Writing - Review \& Editing, Project administration.

\section*{Acknowledgements}   
This study was supported by the Natural Science Foundation of Shandong Province, China (No.ZR2024QF057).

\bibliographystyle{model1-num-names}

\begin{thebibliography}{38}
\expandafter\ifx\csname natexlab\endcsname\relax\def\natexlab#1{#1}\fi
\providecommand{\url}[1]{\texttt{#1}}
\providecommand{\href}[2]{#2}
\providecommand{\path}[1]{#1}
\providecommand{\DOIprefix}{doi:}
\providecommand{\ArXivprefix}{arXiv:}
\providecommand{\URLprefix}{URL: }
\providecommand{\Pubmedprefix}{pmid:}
\providecommand{\doi}[1]{\href{http://dx.doi.org/#1}{\path{#1}}}
\providecommand{\Pubmed}[1]{\href{pmid:#1}{\path{#1}}}
\providecommand{\bibinfo}[2]{#2}
\ifx\xfnm\relax \def\xfnm[#1]{\unskip,\space#1}\fi
\bibitem[{Pensoneault and Zhu(2024)}]{pensoneault2024efficient}
\bibinfo{author}{A.~Pensoneault}, \bibinfo{author}{X.~Zhu},
\newblock \bibinfo{title}{Efficient bayesian physics informed neural networks
  for inverse problems via ensemble kalman inversion},
\newblock \bibinfo{journal}{Journal of Computational Physics}
  \bibinfo{volume}{508} (\bibinfo{year}{2024}) \bibinfo{pages}{113006}.
\bibitem[{Engl et~al.(1996)Engl, Hanke, and Neubauer}]{engl1996regularization}
\bibinfo{author}{H.~W. Engl}, \bibinfo{author}{M.~Hanke},
  \bibinfo{author}{A.~Neubauer}, \bibinfo{title}{Regularization of inverse
  problems}, volume \bibinfo{volume}{375}, \bibinfo{publisher}{Springer Science
  \& Business Media}, \bibinfo{year}{1996}.
\bibitem[{Beck and Arnold(1977)}]{beck1977parameter}
\bibinfo{author}{J.~V. Beck}, \bibinfo{author}{K.~J. Arnold},
  \bibinfo{title}{Parameter estimation in engineering and science},
  \bibinfo{publisher}{James Beck}, \bibinfo{year}{1977}.
\bibitem[{Woodbury(2002)}]{woodbury2002inverse}
\bibinfo{author}{K.~A. Woodbury}, \bibinfo{title}{Inverse engineering
  handbook}, \bibinfo{publisher}{Crc press}, \bibinfo{year}{2002}.
\bibitem[{Wang and Zabaras(2004)}]{wang2004hierarchical}
\bibinfo{author}{J.~Wang}, \bibinfo{author}{N.~Zabaras},
\newblock \bibinfo{title}{Hierarchical bayesian models for inverse problems in
  heat conduction},
\newblock \bibinfo{journal}{Inverse Problems} \bibinfo{volume}{21}
  (\bibinfo{year}{2004}) \bibinfo{pages}{183}.
\bibitem[{Yang et~al.(2021)Yang, Meng, and Karniadakis}]{yang2021b}
\bibinfo{author}{L.~Yang}, \bibinfo{author}{X.~Meng}, \bibinfo{author}{G.~E.
  Karniadakis},
\newblock \bibinfo{title}{B-pinns: Bayesian physics-informed neural networks
  for forward and inverse pde problems with noisy data},
\newblock \bibinfo{journal}{Journal of Computational Physics}
  \bibinfo{volume}{425} (\bibinfo{year}{2021}) \bibinfo{pages}{109913}.
\bibitem[{Yang and Foster(2022)}]{yang2022multi}
\bibinfo{author}{M.~Yang}, \bibinfo{author}{J.~T. Foster},
\newblock \bibinfo{title}{Multi-output physics-informed neural networks for
  forward and inverse pde problems with uncertainties},
\newblock \bibinfo{journal}{Computer Methods in Applied Mechanics and
  Engineering} \bibinfo{volume}{402} (\bibinfo{year}{2022})
  \bibinfo{pages}{115041}.
\bibitem[{Linka et~al.(2022)Linka, Sch{\"a}fer, Meng, Zou, Karniadakis, and
  Kuhl}]{linka2022bayesian}
\bibinfo{author}{K.~Linka}, \bibinfo{author}{A.~Sch{\"a}fer},
  \bibinfo{author}{X.~Meng}, \bibinfo{author}{Z.~Zou}, \bibinfo{author}{G.~E.
  Karniadakis}, \bibinfo{author}{E.~Kuhl},
\newblock \bibinfo{title}{Bayesian physics informed neural networks for
  real-world nonlinear dynamical systems},
\newblock \bibinfo{journal}{Computer Methods in Applied Mechanics and
  Engineering} \bibinfo{volume}{402} (\bibinfo{year}{2022})
  \bibinfo{pages}{115346}.
\bibitem[{Li and Marzouk(2014)}]{li2014adaptive}
\bibinfo{author}{J.~Li}, \bibinfo{author}{Y.~M. Marzouk},
\newblock \bibinfo{title}{Adaptive construction of surrogates for the bayesian
  solution of inverse problems},
\newblock \bibinfo{journal}{SIAM Journal on Scientific Computing}
  \bibinfo{volume}{36} (\bibinfo{year}{2014}) \bibinfo{pages}{A1163--A1186}.
\bibitem[{Neal(2012)}]{neal2012bayesian}
\bibinfo{author}{R.~M. Neal}, \bibinfo{title}{Bayesian learning for neural
  networks}, volume \bibinfo{volume}{118}, \bibinfo{publisher}{Springer Science
  \& Business Media}, \bibinfo{year}{2012}.
\bibitem[{Lee et~al.(2017)Lee, Bahri, Novak, Schoenholz, Pennington, and
  Sohl-Dickstein}]{lee2017deep}
\bibinfo{author}{J.~Lee}, \bibinfo{author}{Y.~Bahri},
  \bibinfo{author}{R.~Novak}, \bibinfo{author}{S.~S. Schoenholz},
  \bibinfo{author}{J.~Pennington}, \bibinfo{author}{J.~Sohl-Dickstein},
\newblock \bibinfo{title}{Deep neural networks as gaussian processes},
\newblock \bibinfo{journal}{arXiv preprint arXiv:1711.00165}
  (\bibinfo{year}{2017}).
\bibitem[{Betancourt(2019)}]{betancourt2019convergence}
\bibinfo{author}{M.~Betancourt},
\newblock \bibinfo{title}{The convergence of markov chain monte carlo methods:
  from the metropolis method to hamiltonian monte carlo},
\newblock \bibinfo{journal}{Annalen der Physik} \bibinfo{volume}{531}
  (\bibinfo{year}{2019}) \bibinfo{pages}{1700214}.
\bibitem[{Neal et~al.(2011)}]{neal2011mcmc}
\bibinfo{author}{R.~M. Neal}, et~al.,
\newblock \bibinfo{title}{Mcmc using hamiltonian dynamics},
\newblock \bibinfo{journal}{Handbook of markov chain monte carlo}
  \bibinfo{volume}{2} (\bibinfo{year}{2011}) \bibinfo{pages}{2}.
\bibitem[{Betancourt(2017)}]{betancourt2017conceptual}
\bibinfo{author}{M.~Betancourt},
\newblock \bibinfo{title}{A conceptual introduction to hamiltonian monte
  carlo},
\newblock \bibinfo{journal}{arXiv preprint arXiv:1701.02434}
  (\bibinfo{year}{2017}).
\bibitem[{Blei et~al.(2017)Blei, Kucukelbir, and
  McAuliffe}]{blei2017variational}
\bibinfo{author}{D.~M. Blei}, \bibinfo{author}{A.~Kucukelbir},
  \bibinfo{author}{J.~D. McAuliffe},
\newblock \bibinfo{title}{Variational inference: A review for statisticians},
\newblock \bibinfo{journal}{Journal of the American statistical Association}
  \bibinfo{volume}{112} (\bibinfo{year}{2017}) \bibinfo{pages}{859--877}.
\bibitem[{Foong et~al.(2019)Foong, Li, Hern{\'a}ndez-Lobato, and
  Turner}]{foong2019between}
\bibinfo{author}{A.~Y. Foong}, \bibinfo{author}{Y.~Li}, \bibinfo{author}{J.~M.
  Hern{\'a}ndez-Lobato}, \bibinfo{author}{R.~E. Turner},
\newblock \bibinfo{title}{{'In-Between'Uncertainty in Bayesian Neural
  Networks}},
\newblock \bibinfo{journal}{arXiv preprint arXiv:1906.11537}
  (\bibinfo{year}{2019}).
\bibitem[{Yao et~al.(2019)Yao, Pan, Ghosh, and Doshi-Velez}]{yao2019quality}
\bibinfo{author}{J.~Yao}, \bibinfo{author}{W.~Pan}, \bibinfo{author}{S.~Ghosh},
  \bibinfo{author}{F.~Doshi-Velez},
\newblock \bibinfo{title}{{Quality of uncertainty quantification for Bayesian
  neural network inference}},
\newblock \bibinfo{journal}{arXiv preprint arXiv:1906.09686}
  (\bibinfo{year}{2019}).
\bibitem[{Ceccarelli(2021)}]{ceccarelli2019bayesian}
\bibinfo{author}{D.~Ceccarelli},
\newblock \bibinfo{title}{Bayesian physics-informed neural networks for inverse
  uncertainty quantification problems in cardiac electrophysiology},
\newblock \bibinfo{journal}{Milan, Italy:Polytechnic Univ. Milan}
  (\bibinfo{year}{2021}).
\bibitem[{Li et~al.(2024)Li, Grana, and Liu}]{li2024bayesian}
\bibinfo{author}{P.~Li}, \bibinfo{author}{D.~Grana}, \bibinfo{author}{M.~Liu},
\newblock \bibinfo{title}{Bayesian neural network and bayesian physics-informed
  neural network via variational inference for seismic petrophysical
  inversion},
\newblock \bibinfo{journal}{Geophysics} \bibinfo{volume}{89}
  (\bibinfo{year}{2024}) \bibinfo{pages}{1--46}.
\bibitem[{Antil et~al.(2021)Antil, Elman, Onwunta, and Verma}]{antil2021novel}
\bibinfo{author}{H.~Antil}, \bibinfo{author}{H.~C. Elman},
  \bibinfo{author}{A.~Onwunta}, \bibinfo{author}{D.~Verma},
\newblock \bibinfo{title}{Novel deep neural networks for solving bayesian
  statistical inverse},
\newblock \bibinfo{journal}{arXiv preprint arXiv:2102.03974}
  (\bibinfo{year}{2021}).
\bibitem[{Lin et~al.(2022)Lin, Wang, and Zhang}]{lin2022multi}
\bibinfo{author}{G.~Lin}, \bibinfo{author}{Y.~Wang},
  \bibinfo{author}{Z.~Zhang},
\newblock \bibinfo{title}{Multi-variance replica exchange sgmcmc for inverse
  and forward problems via bayesian pinn},
\newblock \bibinfo{journal}{Journal of Computational Physics}
  \bibinfo{volume}{460} (\bibinfo{year}{2022}) \bibinfo{pages}{111173}.
\bibitem[{Li et~al.(2023)Li, Wang, and Yan}]{li2023surrogate}
\bibinfo{author}{Y.~Li}, \bibinfo{author}{Y.~Wang}, \bibinfo{author}{L.~Yan},
\newblock \bibinfo{title}{Surrogate modeling for bayesian inverse problems
  based on physics-informed neural networks},
\newblock \bibinfo{journal}{Journal of Computational Physics}
  \bibinfo{volume}{475} (\bibinfo{year}{2023}) \bibinfo{pages}{111841}.
\bibitem[{Sun and Wang(2020)}]{sun2020physics}
\bibinfo{author}{L.~Sun}, \bibinfo{author}{J.-X. Wang},
\newblock \bibinfo{title}{Physics-constrained bayesian neural network for fluid
  flow reconstruction with sparse and noisy data},
\newblock \bibinfo{journal}{Theoretical and Applied Mechanics Letters}
  \bibinfo{volume}{10} (\bibinfo{year}{2020}) \bibinfo{pages}{161--169}.
\bibitem[{Jiang et~al.(2022)Jiang, Wanga, Wena, Li, and Wang}]{jiang2022pinn}
\bibinfo{author}{X.~Jiang}, \bibinfo{author}{X.~Wanga},
  \bibinfo{author}{Z.~Wena}, \bibinfo{author}{E.~Li},
  \bibinfo{author}{H.~Wang},
\newblock \bibinfo{title}{{An E-PINN assisted practical uncertainty
  quantification for inverse problems}},
\newblock \bibinfo{journal}{arXiv preprint arXiv:2209.10195}
  (\bibinfo{year}{2022}).
\bibitem[{Iglesias et~al.(2013)Iglesias, Law, and
  Stuart}]{iglesias2013ensemble}
\bibinfo{author}{M.~A. Iglesias}, \bibinfo{author}{K.~J. Law},
  \bibinfo{author}{A.~M. Stuart},
\newblock \bibinfo{title}{Ensemble kalman methods for inverse problems},
\newblock \bibinfo{journal}{Inverse Problems} \bibinfo{volume}{29}
  (\bibinfo{year}{2013}) \bibinfo{pages}{045001}.
\bibitem[{Meng et~al.(2022)Meng, Yang, Mao, del {\'A}guila~Ferrandis, and
  Karniadakis}]{meng2022learning}
\bibinfo{author}{X.~Meng}, \bibinfo{author}{L.~Yang}, \bibinfo{author}{Z.~Mao},
  \bibinfo{author}{J.~del {\'A}guila~Ferrandis}, \bibinfo{author}{G.~E.
  Karniadakis},
\newblock \bibinfo{title}{Learning functional priors and posteriors from data
  and physics},
\newblock \bibinfo{journal}{Journal of Computational Physics}
  \bibinfo{volume}{457} (\bibinfo{year}{2022}) \bibinfo{pages}{111073}.
\bibitem[{Perez et~al.(2023)Perez, Maddu, Sbalzarini, and
  Poncet}]{perez2023adaptive}
\bibinfo{author}{S.~Perez}, \bibinfo{author}{S.~Maddu}, \bibinfo{author}{I.~F.
  Sbalzarini}, \bibinfo{author}{P.~Poncet},
\newblock \bibinfo{title}{Adaptive weighting of bayesian physics informed
  neural networks for multitask and multiscale forward and inverse problems},
\newblock \bibinfo{journal}{Journal of Computational Physics}
  \bibinfo{volume}{491} (\bibinfo{year}{2023}) \bibinfo{pages}{112342}.
\bibitem[{Xu et~al.(2020)Xu, Zhang, Luo, Xiao, and Ma}]{Xu_2020}
\bibinfo{author}{Z.-Q.~J. Xu}, \bibinfo{author}{Y.~Zhang},
  \bibinfo{author}{T.~Luo}, \bibinfo{author}{Y.~Xiao}, \bibinfo{author}{Z.~Ma},
\newblock \bibinfo{title}{{Frequency principle: Fourier analysis sheds light on
  deep neural networks}},
\newblock \bibinfo{journal}{Communications in Computational Physics}
  \bibinfo{volume}{28} (\bibinfo{year}{2020}) \bibinfo{pages}{1746--1767}.
\bibitem[{Rahaman et~al.(2019)Rahaman, Arpit, Baratin, Draxler, Lin, Hamprecht,
  Bengio, and Courville}]{rahaman2018spectral}
\bibinfo{author}{N.~Rahaman}, \bibinfo{author}{D.~Arpit},
  \bibinfo{author}{A.~Baratin}, \bibinfo{author}{F.~Draxler},
  \bibinfo{author}{M.~Lin}, \bibinfo{author}{F.~A. Hamprecht},
  \bibinfo{author}{Y.~Bengio}, \bibinfo{author}{A.~Courville},
\newblock \bibinfo{title}{On the spectral bias of deep neural networks},
\newblock \bibinfo{journal}{International Conference on Machine Learning}
  (\bibinfo{year}{2019}).
\bibitem[{Wang et~al.(2021)Wang, Wang, and Perdikaris}]{wang2020eigenvector}
\bibinfo{author}{S.~Wang}, \bibinfo{author}{H.~Wang},
  \bibinfo{author}{P.~Perdikaris},
\newblock \bibinfo{title}{{On the eigenvector bias of Fourier feature networks:
  From regression to solving multi-scale PDEs with physics-informed neural
  networks}},
\newblock \bibinfo{journal}{Computer Methods in Applied Mechanics and
  Engineering} \bibinfo{volume}{384} (\bibinfo{year}{2021})
  \bibinfo{pages}{113938}.
\bibitem[{Tancik et~al.(2020)Tancik, Srinivasan, Mildenhall, Fridovich-Keil,
  Raghavan, Singhal, Ramamoorthi, Barron, and Ng}]{tancik2020fourier}
\bibinfo{author}{M.~Tancik}, \bibinfo{author}{P.~Srinivasan},
  \bibinfo{author}{B.~Mildenhall}, \bibinfo{author}{S.~Fridovich-Keil},
  \bibinfo{author}{N.~Raghavan}, \bibinfo{author}{U.~Singhal},
  \bibinfo{author}{R.~Ramamoorthi}, \bibinfo{author}{J.~Barron},
  \bibinfo{author}{R.~Ng},
\newblock \bibinfo{title}{{Fourier features let networks learn high frequency
  functions in low dimensional domains}},
\newblock \bibinfo{journal}{Advances in Neural Information Processing Systems}
  \bibinfo{volume}{33} (\bibinfo{year}{2020}) \bibinfo{pages}{7537--7547}.
\bibitem[{Li et~al.(2023)Li, Xia, Liu, and Liao}]{li2023deep}
\bibinfo{author}{S.~Li}, \bibinfo{author}{Y.~Xia}, \bibinfo{author}{Y.~Liu},
  \bibinfo{author}{Q.~Liao},
\newblock \bibinfo{title}{{A deep domain decomposition method based on Fourier
  features}},
\newblock \bibinfo{journal}{Journal of Computational and Applied Mathematics}
  \bibinfo{volume}{423} (\bibinfo{year}{2023}) \bibinfo{pages}{114963}.
\bibitem[{Zong et~al.(2025)Zong, Barajas-Solano, and
  Tartakovsky}]{zong2025randomized}
\bibinfo{author}{Y.~Zong}, \bibinfo{author}{D.~Barajas-Solano},
  \bibinfo{author}{A.~M. Tartakovsky},
\newblock \bibinfo{title}{Randomized physics-informed neural networks for
  bayesian data assimilation},
\newblock \bibinfo{journal}{Computer Methods in Applied Mechanics and
  Engineering} \bibinfo{volume}{436} (\bibinfo{year}{2025})
  \bibinfo{pages}{117670}.
\bibitem[{Liu et~al.(2020)Liu, Cai, and Xu}]{liu2020multi}
\bibinfo{author}{Z.~Liu}, \bibinfo{author}{W.~Cai}, \bibinfo{author}{Z.-Q.~J.
  Xu},
\newblock \bibinfo{title}{{Multi-scale Deep Neural Network (MscaleDNN) for
  Solving Poisson-Boltzmann Equation in Complex Domains}},
\newblock \bibinfo{journal}{Communications in Computational Physics}
  \bibinfo{volume}{28} (\bibinfo{year}{2020}) \bibinfo{pages}{1970--2001}.
\bibitem[{{Wang} et~al.(2020){Wang}, {Zhang}, and {Cai}}]{wang2020multiscale}
\bibinfo{author}{B.~{Wang}}, \bibinfo{author}{W.~{Zhang}},
  \bibinfo{author}{W.~{Cai}},
\newblock \bibinfo{title}{{Multi-Scale Deep Neural Network (MscaleDNN) Methods
  for Oscillatory Stokes Flows in Complex Domains}},
\newblock \bibinfo{journal}{Communications in Computational Physics}
  \bibinfo{volume}{28} (\bibinfo{year}{2020}) \bibinfo{pages}{2139--2157}.
\bibitem[{Wang et~al.(2024)Wang, Yao, Guo, and Gao}]{wang2024practical}
\bibinfo{author}{Y.~Wang}, \bibinfo{author}{Y.~Yao}, \bibinfo{author}{J.~Guo},
  \bibinfo{author}{Z.~Gao},
\newblock \bibinfo{title}{A practical pinn framework for multi-scale problems
  with multi-magnitude loss terms},
\newblock \bibinfo{journal}{Journal of Computational Physics}
  \bibinfo{volume}{510} (\bibinfo{year}{2024}) \bibinfo{pages}{113112}.
\bibitem[{Li et~al.(2023)Li, Xu, and Zhang}]{li2023subspace}
\bibinfo{author}{X.-A. Li}, \bibinfo{author}{Z.-Q.~J. Xu},
  \bibinfo{author}{L.~Zhang},
\newblock \bibinfo{title}{{Subspace decomposition based DNN algorithm for
  elliptic type multi-scale PDEs}},
\newblock \bibinfo{journal}{Journal of Computational Physics}
  (\bibinfo{year}{2023}) \bibinfo{pages}{112242}.
\bibitem[{John~Lu(2010)}]{john2010elements}
\bibinfo{author}{Z.~John~Lu}, \bibinfo{title}{The elements of statistical
  learning: data mining, inference, and prediction}, \bibinfo{year}{2010}.

\end{thebibliography}

\section*{Appendix A: Hamiltonian Monte Carlo (HMC)}\label{App_HMC}
Hamiltonian Monte Carlo (HMC) augment the parameter space $\bm{\theta}$ into $(\bm{\theta}, \bm{r})$ with an auxiliary momentum variable $\bm{r}$, which matches the dimensionality of $\bm{\theta}$. This formulation enables the use of Hamiltonian dynamics to propose new states in the phase space, facilitating efficient exploration of the posterior distribution. The Hamiltonian $H(\bm{\theta}, \bm{r})$ is composed of the potential energy $U(\bm{\theta})$, derived from the negative log-posterior, and the kinetic energy $K(\bm{r})$, often modeled as $\bm{r}^\top M^{-1} \bm{r} / 2$, where $M$ is the mass matrix. By leveraging the continuous evolution of the Hamiltonian system, HMC avoids random-walk behavior and achieves faster convergence compared to traditional sampling methods.

The process begins by sampling an initial momentum $\bm{r}_0 \sim \mathcal{N}(0, M)$ and setting the initial states $(\bm{\theta}_0, \bm{r}_0)$ to the previous iteration's values $(\bm{\theta}^{t^{k-1}}, \bm{r}^{t^{k-1}})$. To simulate the Hamiltonian dynamics, the leapfrog integrator is employed for $L$ steps with a fixed step size $d$. The updates for momentum and parameters at each step follow these equations:
\[
\begin{aligned}
    \bm{r}_i &\gets \bm{r}_i - \frac{d}{2} \nabla U(\bm{\theta}_i), \\
    \bm{\theta}_{i+1} &\gets \bm{\theta}_i + d M^{-1} \bm{r}_i, \\
    \bm{r}_{i+1} &\gets \bm{r}_i - \frac{d}{2} \nabla U(\bm{\theta}_{i+1}).
\end{aligned}
\]

After completing $L$ leapfrog steps, the Metropolis-Hastings acceptance criterion is applied. The acceptance probability $\alpha$ is given by:
\[
\alpha = \min\left\{1, \exp(H(\bm{\theta}_L, \bm{r}_L) - H(\bm{\theta}^{t^{k-1}}, \bm{r}^{t^{k-1}}))\right\}.
\]

$\bm{\theta}^{L}$ is accepted to be $\bm{\theta}^{t^{k}}$ with probability $\alpha$; otherwise, the previous state $\bm{\theta}^{t^{k-1}}$ is retained. The last few samples will be used for posterior inferences, which is a hyperparameter for user's choice.

The implementation code for HMC can be found at  \href{https://github.com/AdamCobb/hamiltorch}{https://github.com/AdamCobb/hamiltorch}, and the corresponding codes for estimating the parameters of a given neural network are as follows.

\begin{lstlisting}
    def params_grad(para):
        para = para.detach().requires_grad_()
        log_prob = log_prob_func(para)
        para = collect_gradients(log_prob, para, pass_grad)
        if torch.cuda.is_available():
            torch.cuda.empty_cache()
        return para.grad
    ret_params = []
    ret_momenta = []
    momentum += 0.5 * step_size * params_grad(params)
    for n in range(steps):
        if inv_mass is None:
            params = params + step_size * momentum 
        else:
            if type(inv_mass) is list:
                i = 0
                for block in inv_mass:
                    it = block[0].shape[0]
                    params[i:it+i] = params[i:it+i] + step_size * torch.matmul(block,momentum[i:it+i].view(-1,1)).view(-1)
                    i += it
            elif len(inv_mass.shape) == 2:
                params = params + step_size * torch.matmul(inv_mass,momentum.view(-1,1)).view(-1)
            else:
                params = params + step_size * inv_mass * momentum
        p_grad = params_grad(params)
        momentum += step_size * p_grad
        ret_params.append(params.clone())
        ret_momenta.append(momentum.clone())
    ret_momenta[-1] = ret_momenta[-1] - 0.5 * step_size * p_grad.clone()
\end{lstlisting}

\begin{lstlisting}
    def log_prob_func(params):
        params_unflattened = util.unflatten(model, params)
        i_prev = 0
        l_prior = torch.zeros_like( params[0], requires_grad=True) 
        for weights, index, shape, dist in zip(model.parameters(), params_flattened_list, params_shape_list, dist_list):
            w = params[i_prev:index+i_prev]
            l_prior = dist.log_prob(w).sum() + l_prior
            i_prev += index

        if x is None:
            return l_prior/prior_scale

        x_device = x.to(device)
        y_device = y.to(device)

        output = fmodel(x_device, params=params_unflattened)

        if model_loss == 'binary_class_linear_output':
            crit = nn.BCEWithLogitsLoss(reduction='sum')
            ll = - tau_out *(crit(output, y_device))
        elif model_loss == 'multi_class_linear_output':
            crit = nn.CrossEntropyLoss(reduction='sum')
            ll = - tau_out *(crit(output, y_device.long().view(-1)))
        elif model_loss == 'multi_class_log_softmax_output':
            ll = - tau_out *(torch.nn.functional.nll_loss(output, y_device.long().view(-1)))
        elif model_loss == 'regression':
            ll = - 0.5 * tau_out * ((output - y_device) ** 2).sum(0)

        elif callable(model_loss):
            ll = - model_loss(output, y_device).sum(0)
        else:
            raise NotImplementedError()

        if torch.cuda.is_available():
            del x_device, y_device
            torch.cuda.empty_cache()

        if predict:
            return (ll + l_prior/prior_scale), output
        else:
            return (ll + l_prior/prior_scale)
\end{lstlisting}

\end{document}